\documentclass{article}

 \usepackage[main, final]{neurips_2026}

\usepackage[utf8]{inputenc}
\usepackage[T1]{fontenc}
\usepackage{url}
\usepackage{booktabs}
\usepackage{amsfonts}
\usepackage{nicefrac}
\usepackage{microtype}
\usepackage[table]{xcolor}
\usepackage{xspace}
\definecolor{ceruleanblue}{rgb}{0.16, 0.32, 0.75}
\usepackage[pagebackref,breaklinks,colorlinks,citecolor=ceruleanblue, linkcolor=ceruleanblue]{hyperref}
\usepackage{graphicx}
\usepackage{amsmath}
\usepackage[capitalize]{cleveref}
\crefname{section}{Sec.}{Secs.}
\Crefname{section}{Section}{Sections}
\Crefname{table}{Table}{Tables}
\crefname{table}{Tab.}{Tabs.}
\usepackage[toc,page]{appendix}
\crefname{appendix}{Appendix}{Appendices}
\Crefname{appendix}{Appendix}{Appendices}
\usepackage{wrapfig}

\usepackage{enumitem}

\usepackage{multirow}
\newcommand{\methodName}{RefineAny3D\xspace}
\newcommand{\ourMethod}{RefineAny3D\xspace}
\newcommand{\benchmarkName}{Refine3D\xspace}
\newcommand{\methodNameFull}{RefineAny3D: Depth Refinement as Semantic Alignment for Monocular 3D Detection}

\definecolor{car}{RGB}{230,25,75}
\definecolor{cyclist}{RGB}{255, 130, 48}
\definecolor{pedestrian}{RGB}{138, 43, 226}
\definecolor{colorgt}{RGB}{170,255,195}
\definecolor{white}{rgb}{1.0, 1.0, 1.0}
\definecolor{monocop}{RGB}{71, 159, 179}
\definecolor{baseline}{RGB}{253, 190, 110}
\definecolor{lightgray}{gray}{0.95}
\definecolor{mygreen}{RGB}{180,215,195}

\definecolor{mylightblue}{RGB}{170,200,235}

\definecolor{mypeach}{RGB}{238,200,164}

\newcommand{\monoThreeD}{Mono3D\xspace}

\newcommand{\twoD}{2D\xspace}
\newcommand{\threeD}{3D\xspace}

\newcommand{\kitti}{KITTI\xspace}
\newcommand{\nuscenes}{nuScenes\xspace}

\newcommand{\val}{Val\xspace}

\newcommand{\ap}{AP}

\newcommand{\apThreeD}{\ap$_{3\text{D}}$\xspace}

\newcommand{\monodetr}{MonoDETR\xspace}
\newcommand{\monodgp}{MonoDGP\xspace}

\newcommand{\fdThreeD}{FD3D\xspace}

\newcommand{\monocd}{MonoCD\xspace}
\newcommand{\monomae}{MonoMAE\xspace}

\newcommand{\omniThreeD}{Omni3D\xspace}

\newcommand{\monotakd}{MonoTAKD\xspace}

\newcommand{\ovMonoThreeD}{OVMono3D\xspace}

\newcommand{\eg}{\emph{e.g.}\xspace}

\usepackage{pifont}
\newcommand{\noIndentHeading}[1]{\noindent\textbf{#1}}

\definecolor{XLcolor}{rgb}{0.858, 0.188, 0.478}

\title{\methodNameFull}

\author{%
  Zhihao Zhang\textsuperscript{1}\quad Gengwei Zhang\textsuperscript{2} \quad Tianlong Chen\textsuperscript{2} \quad Xiaoming Liu\textsuperscript{1,2} \\
\textsuperscript{1}Michigan State University \quad \textsuperscript{2}University of North Carolina at Chapel Hill
}

\begin{document}

\maketitle

\begin{abstract}
Monocular \threeD object detection spans two regimes: closed-set detectors operating within a fixed category vocabulary, and open-vocabulary detectors that localize arbitrary categories by leveraging depth foundation models for \threeD geometry. We find that current depth foundation models, despite their strong zero-shot generalization, lack the object-level precision \threeD detection demands: substituting a state-of-the-art depth foundation model for a strong detector's predicted depth degrades accuracy.
Rather than pushing detectors or depth models to be more accurate end-to-end, we treat object-level depth refinement as a stand-alone task and present \ourMethod, a vision-language model that corrects depth without ever predicting a numerical value. Our key insight is that depth error has a direct visual signature in image space: when projected onto the image, a correctly placed box tightly encloses the object, while a too-far box projects too small and a too-close box projects too large. Depth refinement thus reduces to a visual alignment problem rather than a metric regression problem, which we instantiate by extending the VLM's vocabulary with action tokens that replace numerical depth output with categorical decisions, and by supervising the model on a large-scale chain-of-thought dataset that grounds each decision in explicit visual evidence.
Applied as a single post-hoc step, \ourMethod delivers consistent gains across closed-set detectors, open-vocabulary detectors, and 3D auto-labeling tools, and generalizes to novel categories, scenes, and cameras without retraining. Code is available at \url{https://github.com/alanzhangcs/RefineAny3D}.
\end{abstract}

\section{Introduction}
\label{sec:intro}

Monocular \threeD object detection (\monoThreeD), which aims to localize and recognize  objects in \threeD space from a single RGB image, is a fundamental problem with broad applications in autonomous driving~\cite{chen2024end, chen2016monocular}, robotics~\cite{wang2024embodiedscan, zhu2014single}, and augmented reality~\cite{liu2019edge}. Compared to LiDAR-, radar-, or multi-view-based alternatives~\cite{yin2021center, shi2019pointrcnn, long2023radiant, long2025riccardo, peng2024learning}, \monoThreeD is more cost-effective and scalable, yet the absence of explicit depth cues makes accurate \threeD reasoning inherently challenging. Existing \monoThreeD detectors broadly fall  into two regimes. \emph{Closed-set} detectors~\cite{MonoRCNN_ICCV21, zhang2023monodetr} are trained and evaluated on a fixed category vocabulary within a single dataset; while accurate in-domain, they cannot handle novel categories. \emph{Open-vocabulary} \threeD detectors~\cite{yao2025open, Yang_2025_ICCV, zhang2025detect}, in contrast, aim to recognize and localize arbitrary categories in unseen domains, a setting where category-specific geometric priors no longer apply and depth must be recovered from generic visual cues alone.

\begin{figure}[t]
    \centering
    \includegraphics[width=\linewidth]{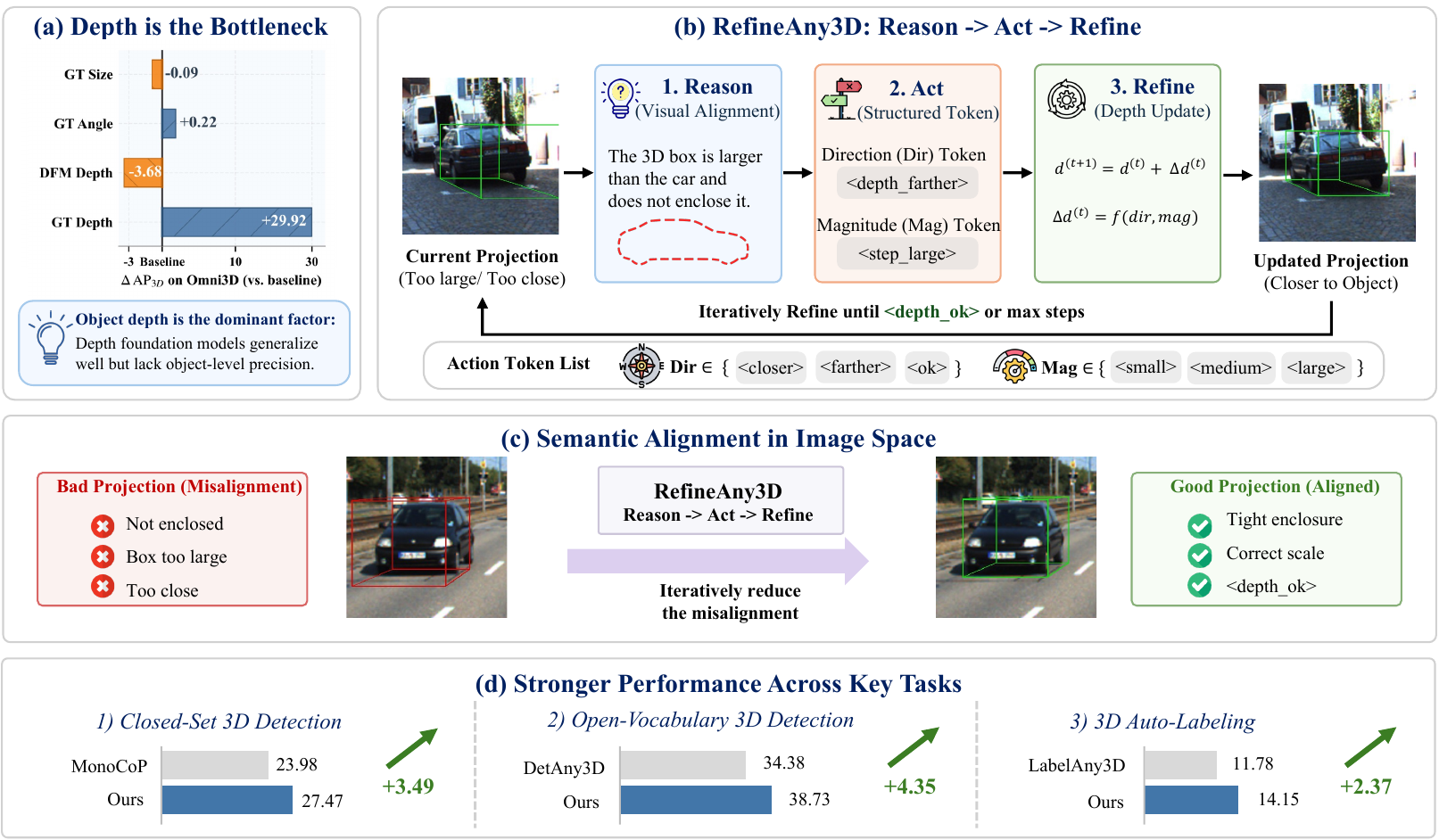}
    \small
    \caption{
        \textbf{\ourMethod: \threeD box refinement as semantic alignment in image space.}
        \textbf{(a)} Object depth is the dominant bottleneck of monocular \threeD detection: replacing predicted depth with ground truth boosts \apThreeD by $29.92$, while substituting a depth foundation model (DFM) degrades it by $3.68$. Relying on DFM alone is insufficient for accurate object-level depth.
        \textbf{(b)} \ourMethod reformulates depth refinement as a semantic alignment problem in image space. Given a candidate \threeD box projected onto the image, a vision-language model emits action tokens that iteratively adjust the depth until the projection aligns with the target or a step limit is reached.
        \textbf{(c)} Visual signature of depth error: a too-close box projects too large (left), while a correctly placed box tightly encloses the target object (right).
        \textbf{(d)} As a plug-and-play module, \ourMethod consistently improves closed-set detectors, open-vocabulary detectors, and \threeD auto-labeling methods.
    }
    \label{fig:teaser}
\end{figure}

This need has made depth foundation models~\cite{piccinelli2025unidepthv2, wang2025moge} the de facto source of \threeD geometry for open-vocabulary \monoThreeD, since these models offer strong zero-shot generalization to unseen domains and categories. Yet despite this generalization, their predicted object depth is not precise enough at the level that \threeD IoU thresholds demand: substituting a state-of-the-art depth foundation model~\cite{wang2025moge} for a strong detector's predicted depth in fact degrades \apThreeD by $3.68$ on \omniThreeD~\cite{brazil2023omni3d} (\cref{fig:teaser}; details in \cref{sec:appendix_oracle}). This depth bias is then inherited by every downstream pipeline that builds on these models, including auto-labeling tools~\cite{yao2025labelany3d, wang2025n3d} that distill their predictions into pseudo-\threeD annotations. The object-level geometric precision of open-vocabulary \monoThreeD systems thus remains a fundamental open challenge.

This observation motivates a different approach. Rather than pushing depth foundation models or open-vocabulary \threeD detectors to be more accurate end-to-end, we ask: \emph{can the residual object-depth error of any existing \threeD pipeline be corrected after the fact, by a generic refinement module}? We formalize this as the task of \textbf{\threeD box refinement}: given an image and a candidate \threeD box from any upstream source (a closed-set detector, an open-vocabulary detector, or an auto-labeling tool), output a refined \threeD box whose object depth is closer to the ground truth, leaving other attributes untouched. A successful refinement module is plug-and-play, depends on no specific upstream architecture, and generalizes across diverse object categories, scene types, and camera setups.

We introduce \ourMethod, the first refinement module designed for this task. Our key insight is that the correctness of an object's depth has a direct visual signature: when a candidate \threeD box is projected back onto the image, a correctly placed box tightly encloses the target object, while a too-far box projects too small and a too-close box too large. Refining depth therefore reduces to a familiar visual judgment that depends only on object appearance and \twoD spatial layout, the kind of reasoning vision-language models (VLMs) are well positioned to make. We thus recast depth refinement from a metric regression problem into a semantic alignment problem in image space.

Realizing this idea is not straightforward, since off-the-shelf VLMs~\cite{Qwen2-VL, Qwen2.5-VL, Qwen3-VL} are not trained to reason about projected \threeD boxes or to relate small visual misalignments to depth corrections. We therefore curate a large-scale training dataset on top of \omniThreeD that pairs each candidate box with a chain-of-thought reasoning trace and a corresponding refinement decision, forcing the model to ground its decisions in explicit visual evidence. To make these decisions atomic and unambiguous, we further extend the VLM's vocabulary with a small set of dedicated action tokens, each carrying its own learnable embedding. This sidesteps the VLM's well-known weakness at numerical regression~\cite{lai2024lisa,yang2023lisa++}: the model never emits a continuous depth value, only one of a few discrete decisions, which are then chained iteratively to recover fine-grained precision.

Because refinement depends only on the local alignment between a projected box and its target object, \ourMethod generalizes across categories, scenes, and cameras without retraining. We show that \ourMethod, applied as a plug-and-play module, consistently improves the depth accuracy of closed-set \threeD detectors, open-vocabulary \threeD detectors, and \threeD auto-labeling tools alike, establishing \threeD box refinement as a practical capability that complements existing \monoThreeD systems.

Our contributions are as follows:
\begin{itemize}[leftmargin=*, noitemsep, nolistsep]
\item We formalize \threeD box refinement as a new task that operates on top of any upstream \monoThreeD method, positioning refinement as a capability orthogonal to existing detectors.
\item We recast depth refinement from metric regression into semantic alignment in image space, where a VLM judges depth correctness from visual cues rather than predicting metric depth.

\item We instantiate \ourMethod with action tokens, a small set of dedicated VLM-vocabulary tokens that turn depth refinement into a sequence of categorical decisions, sidestepping the VLM's weakness at numerical regression and enabling iterative recovery of fine-grained precision.

\item Applied as a plug-and-play module, \ourMethod consistently improves closed-set detectors, open-vocabulary detectors, and \threeD auto-labeling tools across novel categories, scenes, and cameras.
\end{itemize}

\section{Related Work}
\label{sec:related}

\noIndentHeading{Monocular 3D Detection.}
\monoThreeD methods can be broadly grouped into closed-set and open-vocabulary settings. Closed-set methods are trained and evaluated on a fixed category vocabulary, typically on benchmarks such as \kitti~\cite{geiger2012we} and \nuscenes~\cite{caesar2020nuscenes}. Early CNN-based detectors adopt center-based prediction~\cite{liu2020smoke, ma2021delving, wang2022probabilistic, liu2023monocular}, exploit \twoD--\threeD geometric consistency~\cite{li2022diversity, lu2021geometry, zhang2021objects, wu2024fd3d, brazil2020kinematic, wang2026span, jeon2026moca3d}, introduce depth-equivariant operators~\cite{kumar2022deviant, qin2022monoground}, build on \twoD detector backbones~\cite{brazil2019m3d, kumar2021groomed, brazil2023omni3d}, lift image features into \threeD voxel grids~\cite{liu2021voxel}, or add bird's-eye-view segmentation supervision~\cite{kumar2024seabird}, while more recent transformer-based detectors~\cite{carion2020end, chen2023group, zhao2024detrs, huang2022monodtr, wu2023monopgc, zhou2023monoatt, zhang2023monodetr, pu2024monodgp, liu2025monotakd, zhang2025unleashing} have come to dominate these benchmarks. Others improve generalization across cameras by conditioning the detector on camera intrinsics~\cite{zhang2026intrinsic} or by making depth robust to unseen camera heights~\cite{kumar2025charm3r}. In parallel, open-vocabulary methods generalize \monoThreeD beyond a fixed category set by scaling \threeD training data or transferring foundation-model priors. Specifically, \omniThreeD~\cite{brazil2023omni3d} unifies existing \threeD detection datasets into a single large-scale benchmark used to train unified detectors~\cite{li2024unimode}, auto-labeling pipelines~\cite{yao2025labelany3d, wang2025n3d, huang2024ovm3d, skvrna2025monosowa, lee2026plot} turn unlabeled images or videos into pseudo-annotated \threeD data via \threeD-aware foundation models~\cite{wang2025moge}, and methods such as \ovMonoThreeD~\cite{yao2025open}, 3D-MOOD~\cite{Yang_2025_ICCV}, and DetAny3D~\cite{zhang2025detect} couple \twoD vision foundation models~\cite{sun2023eva, liu2023grounding} with depth foundation models~\cite{piccinelli2025unidepthv2, wang2025moge} to recognize and localize novel categories in unseen domains~\cite{wang2024ov, detone2026boxer}. Since \ourMethod operates as a detector-agnostic refinement model, it can be applied to both streams, and our experiments show that object-level depth remains an under-addressed bottleneck rather than a detector-specific limitation.

\noIndentHeading{Visual Prompting and Semantic Alignment.}
Recent work shows that multimodal models can benefit from explicit visual cues rendered directly on top of the input image. Early visual prompt engineering with red circles~\cite{shtedritski2023does} demonstrates that simple marks can direct VLM attention to specific regions, while Set-of-Mark prompting~\cite{yang2023set}, ViP-LLaVA~\cite{cai2024vip}, and Draw-and-Understand~\cite{lin2024draw} extend this idea to boxes, masks, points, or arrows that make region references easier for MLLMs to parse. Omni-RGPT~\cite{heo2025omni} and VideoRefer~\cite{yuan2025videorefer} further adapt such region cues to video understanding. These methods primarily use overlaid marks to improve visual understanding, grounding, and referring. Although our \ourMethod follows the broader idea of reasoning over an overlaid visual cue, we use the projected \threeD box as an alignment indicator of how the candidate depth should be refined.

\noIndentHeading{Vision-Language Models for 3D Understanding.}
A growing body of work extends vision-language models (VLMs) toward \threeD understanding. One line equips VLMs with a point-cloud encoder~\cite{3dllm, xu2024pointllm, SpatialLM}, often pre-trained to align \threeD shapes with image and language representations~\cite{zhang2024tamm}; RGB images are lifted into point clouds via a \threeD foundation model~\cite{wang2025vggt} and a point-cloud-aware VLM is trained on top. Another line adopts a dual-backbone design~\cite{cheng2024spatialrgpt, fan2025vlm, qu2026loc3rvlm} that augments the vision encoder with a \threeD or depth foundation model~\cite{piccinelli2025unidepthv2, wang2025moge}, or distills \threeD features into the visual representation~\cite{zhen20243dvla, hu2025g2vlmgeometrygroundedvision, li2025spatial}. A third strengthens the VLM's ability to regress numerical \threeD quantities through reinforcement learning or numerical supervision~\cite{chen2024spatialvlm, huang2025vision, shen2025vlm, huang20253d, qu2026monovlm, cheng2026grounded}. These approaches all task the VLM with directly producing \threeD output, whether raw coordinates or metric depth. \ourMethod takes a different route: we avoid VLM \threeD prediction altogether and instead reduce \threeD refinement to a \twoD semantic alignment problem in image space, solved by emitting a small set of discrete decisions. From another perspective, the use of discrete action outputs also appears in vision-language-action (VLA) models for robotic control~\cite{zitkovich2023rt,kim2024openvla}. Differently, \ourMethod uses action tokens to represent relative depth-refinement decisions inferred from projected-box alignment, and recovers metric precision for monocular 3D detection through iterative updates.

\begin{figure}[t]
    \centering
    \includegraphics[width=\linewidth]{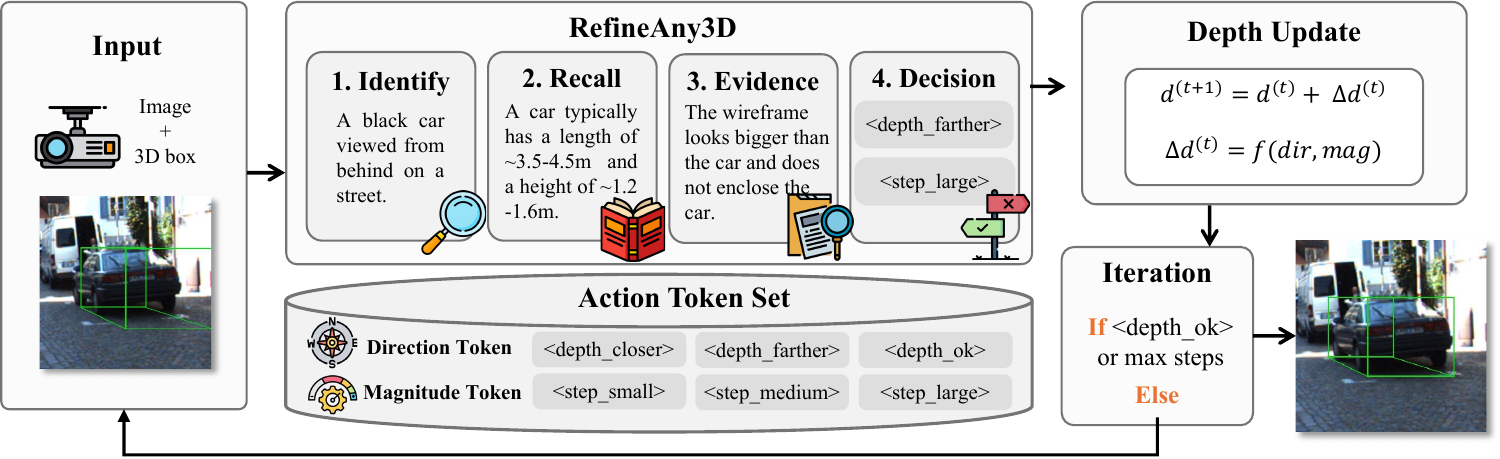}
    \small
    \caption{
        \textbf{Overview of \ourMethod.}
        \ourMethod recasts depth refinement as semantic alignment in image space: when a candidate \threeD box is projected onto the image, depth error has a direct visual signature, with a too-far box projecting too small and a too-close box too large. We render the projected box as a wireframe overlay on the image, and a vision-language model performs chain-of-thought reasoning over the visual misalignment to emit two action tokens specifying the corrective \emph{direction} (closer / ok / farther) and \emph{magnitude} (small / medium / large). These tokens define an object-relative depth update that is applied to the box, and the loop repeats with the updated wireframe until the model emits $\langle\texttt{depth\_ok}\rangle$ or a step cap is reached.
    }
    \label{fig:overview}
\end{figure}

\section{Approach}
\label{sec:approach}

\textbf{Overview.}
\ourMethod refines the object depth of a candidate \threeD box by reformulating depth correction as a \twoD semantic alignment problem (\cref{fig:overview}). Given a noisy box, \ourMethod projects it onto the image, queries a vision-language model to judge how the projection misaligns with the target object, and updates the depth based on a discrete action emitted by the model. The same step is applied iteratively, so that fine-grained corrections emerge from chaining categorical decisions rather than regressing a continuous metric quantity. \cref{sec:formulation} formalizes the alignment view, \cref{sec:action_tokens} introduces the discrete action tokens that realize it, and \cref{sec:training_inference} describes the supervision and inference protocols.

\textbf{Preliminaries: Monocular \threeD Detection.}
A \monoThreeD detector takes an RGB image $\mathbf{I} \in \mathbb{R}^{H \times W \times 3}$ and camera intrinsics $\mathbf{K}$ as input, and outputs a set of \threeD boxes $\{\mathbf{b}_i\}_{i=1}^{N}$. Each box is parameterized as $\mathbf{b} = (x, y, z, w, h, l, \theta)$, where $(x, y, z)$ is the object center in the camera coordinate frame, $(w, h, l)$ are the box dimensions, and $\theta$ is the yaw angle. We write $d = z$ for the object depth, the component of the center along the camera's optical axis. \ourMethod takes such a candidate box $\mathbf{b}$ as input and outputs a refined box $\mathbf{b}'$ with corrected depth.

\subsection{Depth Refinement as Semantic Alignment}
\label{sec:formulation}

Directly regressing a corrected metric depth is difficult: depth is a continuous, scale-sensitive quantity that depends jointly on camera intrinsics, object geometry, and scene context. Our key observation is that once the candidate \threeD box is projected onto the image plane, the camera intrinsics and metric scale are absorbed into the projection itself, and what remains is a purely \twoD question: \emph{does the projected box tightly enclose the target object?} If the projection is too small, the box is too far; if too large, too close. The hard metric problem of correcting depth thereby reduces to a semantic alignment task in image space, judged entirely from \twoD visual evidence. \cref{sec:exp_nonvlm} shows that simpler non-VLM alternatives to this judgment fall short.

To realize this view, we cast depth refinement as a conditional generation problem over a VLM's token sequence. At step $t$, the candidate box $\mathbf{b}^{(t)}$ is projected onto the image and rendered as a wireframe overlay:
\begin{equation}
    \mathbf{I}^{(t)} = \mathrm{render}\!\left(\mathbf{I},\ \pi(\mathbf{b}^{(t)};\,\mathbf{K})\right),
    \label{eq:render}
\end{equation}
where $\pi(\mathbf{b};\mathbf{K})$ projects the eight corners of $\mathbf{b}$ to 2D image coordinates using $\mathbf{K}$. The rendered image $\mathbf{I}^{(t)}$ encodes the current alignment hypothesis: a VLM looking at $\mathbf{I}^{(t)}$ can compare the wireframe with the underlying object directly, without ever computing distances in metric space.

Conditioned on $\mathbf{I}^{(t)}$, the VLM autoregressively generates a single token sequence
\begin{equation}
    \mathcal{S}^{(t)} = \bigl(\mathbf{r}^{(t)},\ \mathbf{a}^{(t)}\bigr),
\end{equation}
where $\mathbf{r}^{(t)}$ is a chain-of-thought reasoning trace and $\mathbf{a}^{(t)}$ is a compact action; the concrete form of $\mathbf{a}^{(t)}$ is given in \cref{sec:action_tokens}. The autoregressive nature of this single generation pass implies the following decomposition by the chain rule:
\begin{equation}
    P\!\left(\mathcal{S}^{(t)} \mid \mathbf{I}^{(t)}\right)
    = \underbrace{P\!\left(\mathbf{r}^{(t)} \mid \mathbf{I}^{(t)}\right)}_{\text{visual reasoning}}\,
      \cdot\,\underbrace{P\!\left(\mathbf{a}^{(t)} \mid \mathbf{I}^{(t)},\, \mathbf{r}^{(t)}\right)}_{\text{action prediction}},
    \label{eq:factorization}
\end{equation}
where the action prediction is conditioned on the reasoning trace produced earlier in the same sequence. This decomposition exposes two complementary design choices that drive the rest of the method: the reasoning prefix $\mathbf{r}^{(t)}$ grounds the action in explicit visual reasoning about the current projection, and the action $\mathbf{a}^{(t)}$ is structured as discrete decisions to sidestep the VLM's known weakness at numerical regression.

\subsection{From Alignment to Action Tokens}
\label{sec:action_tokens}

The formulation in \cref{eq:factorization} reduces depth refinement to producing an action $\mathbf{a}^{(t)}$ at each step. The most direct realization would be to let the VLM verbalize this action in natural language (\eg, ``the box is slightly too far'') or, even more directly, regress a numerical correction $\Delta d$. Both options inherit the VLM's well-known unreliability at fine-grained numerical or open-ended outputs. We instead represent $\mathbf{a}^{(t)}$ with a small set of \emph{action tokens} that encode the action categorically: a direction token $a_d$ specifying how the box should move, paired with a magnitude token $a_m$ specifying how strongly. We extend the VLM's vocabulary with six new special tokens (three for direction, three for magnitude), each carrying its own learnable embedding tuned for depth adjustment, defining a closed output space that simplifies decoding and supervision.

\noindent\textbf{Direction tokens.}
The direction token $a_d \in \{\langle\texttt{depth\_closer}\rangle,\ \langle\texttt{depth\_ok}\rangle,\ \langle\texttt{depth\_farther}\rangle\}$ encodes the qualitative outcome of the alignment check: $\langle\texttt{depth\_closer}\rangle$ when the projected box is too small (the box should be moved closer to the camera, decreasing depth), $\langle\texttt{depth\_farther}\rangle$ when it is too large (the box should be moved farther away, increasing depth), and $\langle\texttt{depth\_ok}\rangle$ when the projection already aligns with the target. We map them to scalar signs by
\begin{equation}
\mathrm{dir}(\langle\texttt{depth\_closer}\rangle) = -1,\quad \mathrm{dir}(\langle\texttt{depth\_ok}\rangle) = 0,\quad \mathrm{dir}(\langle\texttt{depth\_farther}\rangle) = +1.
\end{equation}
Discretizing direction into three classes turns this part of the action into a classification problem that VLMs handle reliably; the precision lost by discretization is recovered through iterative refinement at inference time.

\noindent\textbf{Magnitude tokens.}
The magnitude token $a_m \in \{\langle\texttt{step\_small}\rangle,\allowbreak\ \langle\texttt{step\_medium}\rangle,\allowbreak\ \langle\texttt{step\_large}\rangle\}$ specifies how strongly to adjust along the chosen direction. A fixed metric step size would be inappropriate here: a $0.5\,\text{m}$ shift is negligible for a distant truck but catastrophic for a nearby cup. We therefore tie magnitudes to the object's own characteristic size $s_{\text{obj}} = (w + h + l) / 3$, the arithmetic mean of the box's \threeD dimensions, so that step sizes are \emph{relative} to the object scale:
\begin{equation}
\mathrm{mag}(a_m,\, s_{\text{obj}}) = \alpha_{a_m}\, s_{\text{obj}}, \quad \alpha_{\langle\texttt{step\_small}\rangle} < \alpha_{\langle\texttt{step\_medium}\rangle} < \alpha_{\langle\texttt{step\_large}\rangle}.
\end{equation}
This scale-relative design keeps the same action semantically consistent across objects of different sizes, which is important for generalization across categories and scenes. The full action is then $\mathbf{a}^{(t)} = (a_d^{(t)}, a_m^{(t)})$, and the depth update becomes
\begin{equation}
\Delta d^{(t)} = \mathrm{dir}(a_d^{(t)}) \cdot \mathrm{mag}(a_m^{(t)},\ s_{\text{obj}}).
\end{equation}

\subsection{Training and Inference}
\label{sec:training_inference}

\ourMethod is trained in two stages, both using single-step supervision: given a noisy box paired with its ground-truth chain-of-thought and target action tokens, the model predicts $(a_d, a_m)$ in one pass. In Stage 1, we initialize each new action-token embedding to the mean embedding of a short phrase describing its alignment semantics, freeze the VLM, and train only the input and output embeddings of these six tokens to stabilize them before joint optimization. In Stage 2, we jointly fine-tune the LLM backbone and the action-token embeddings while freezing the vision encoder, preserving the general visual priors from VLM pretraining that we find essential for generalization across novel categories, scenes, and cameras.
At inference, although trained with single-step supervision, \ourMethod refines a box iteratively: at each step we render the current box onto the image, query the VLM for an action, and update the depth, until the model emits $\langle\texttt{depth\_ok}\rangle$ or a safeguard cap of $T_{\max}$ steps is reached. This train-inference asymmetry is justified because each action is a local decision conditioned only on the current visual state, so single-step supervision generalizes naturally to multi-turn refinement.

\section{Data Curation at Scale}
\label{sec:data_curation}

\begin{figure}[t]
    \centering
    \includegraphics[width=\linewidth]{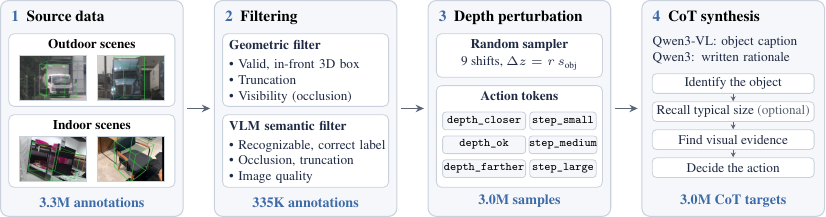}
    \small
    \caption{
        \textbf{Data construction pipeline for training \ourMethod.}
        Scenes are selected, filtered by geometric and VLM-based checks, perturbed in depth with balanced action-token targets, and paired with chain-of-thought rationales; details in \cref{sec:appendix_data}.
    }
    \label{fig:dataset-curation}
\end{figure}

Training \ourMethod requires large-scale triplets of $(\text{image},\ \text{noisy \threeD box},\ \text{refinement target})$, where each target encodes the visual reasoning that maps a perturbed projection back to a corrective action. We curate this data from \omniThreeD~\cite{brazil2023omni3d} through a four-stage pipeline (\cref{fig:dataset-curation}) summarized below; full construction details are in \cref{sec:appendix_data}.

\textbf{Source datasets.}
We build on the \omniThreeD release, which unifies six existing \threeD detection datasets into a single benchmark under a shared camera convention, spanning indoor and outdoor scenes, photographic and synthetic imagery, and fifty object categories. This diversity is essential for \ourMethod's generalization across categories, scenes, and cameras.

\textbf{Annotation filtering.}
The raw release contains many annotations unsuitable as positive training examples, including invalid \threeD boxes, severe occlusion, and mis-categorized labels. We apply a two-stage filter, combining geometric checks on \omniThreeD's metadata fields with a semantic check that queries a VLM to verify object recognizability and label correctness, retaining $335$K annotations.

\textbf{Depth perturbation.}
For each filtered annotation, we synthesize noisy \threeD boxes by perturbing the ground-truth (GT) depth along the camera ray while keeping all other attributes fixed. Perturbation magnitudes are tied to the object's own size $s_{\text{obj}}$ to match the object-relative action design in \cref{sec:action_tokens}, and stratified so that all action-token outcomes are uniformly represented. This yields approximately $3.0$M training samples with an action-token distribution balanced by construction.

\begin{figure}[t]
    \centering
    \includegraphics[width=\linewidth]{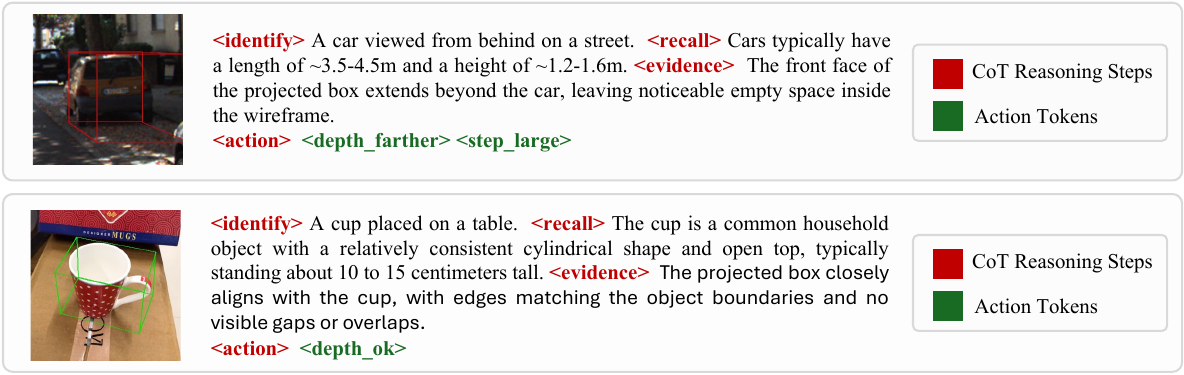}
    \small
    \caption{
        \textbf{Examples of CoT-based supervision.}
        Each training sample pairs a rendered image of a projected \threeD wireframe with a chain-of-thought reasoning target that grounds the corrective action in explicit visual evidence before emitting the action tokens.
    }
    \label{fig:dataset-example}
\end{figure}

\textbf{Chain-of-thought reasoning.}
Supervising only the final action tokens leaves the model free to take shortcuts that do not generalize. We therefore pair each sample with a chain-of-thought (CoT) prefix (\cref{fig:dataset-example}) that the model produces before emitting the action. Each CoT is synthesized by a two-model pipeline: a VLM~\cite{Qwen3-VL} identifies the target object from the image, and a text-only LLM~\cite{qwen3} composes this object identification together with the sample metadata into a reasoning trace consistent with the corrective decision. This forces the model to ground every action in explicit visual evidence rather than memorize input-output shortcuts.

\section{Experiments}
\label{sec:experiments}

We evaluate \ourMethod along four complementary axes: as a standalone refiner on a controlled benchmark (\cref{sec:exp_benchmark}), and as a plug-and-play module on top of three downstream pipelines, namely closed-set \threeD detectors (\cref{sec:exp_closedset}), open-vocabulary \threeD detectors (\cref{sec:exp_openvocab}), and \threeD auto-labeling methods (\cref{sec:exp_autolabel}). We further show that simpler non-VLM mechanisms cannot replace the VLM (\cref{sec:exp_nonvlm}), and verify that fine-tuning for refinement does not erode the underlying VLM's general capability (\cref{sec:exp_cvbench}).

\subsection{\benchmarkName: Refinement as a Standalone Capability}
\label{sec:exp_benchmark}

\benchmarkName is a controlled benchmark we construct on top of \omniThreeD: each test sample pairs a GT \threeD box with a synthetically perturbed noisy version, and the task is to recover the GT. We evaluate on three settings: Standard (in-distribution), Novel Category (open-vocabulary split of \ovMonoThreeD~\cite{yao2025open}), and Novel Camera (rescaled images with updated intrinsics). We report DirAcc (direction accuracy), FullAcc (direction and magnitude both correct), and DepthErr (residual gap between refined and GT depths).

\cref{tab:refine3d} reveals three findings. First, off-the-shelf Qwen3-VL-8B falls below the random-guess baselines ($\sim\!28$ DirAcc, $\sim\!11$ FullAcc), indicating systematic bias rather than uninformative priors. Second, fine-tuning Qwen3-VL with our data using plain-text actions lifts Standard performance substantially ($75.6 / 55.4$), confirming that the supervision signal is informative. Third, \ourMethod outperforms both baselines on all settings and generalizes to both shifts: it retains nearly all accuracy under Novel Camera ($-3.6$ DirAcc, $-3.5$ FullAcc) thanks to its camera-agnostic alignment design, and under Novel Category its FullAcc ($58.5$) still exceeds the plain-text variant.

\begin{table}[t]
    \centering
    \small
    \caption{
        \textbf{Evaluation on \benchmarkName}, our standalone refinement benchmark on \omniThreeD. \ourMethod outperforms both off-the-shelf and fine-tuned Qwen3-VL baselines across all three settings.
    }
    \label{tab:refine3d}
    \resizebox{\linewidth}{!}{
        \tabcolsep=3pt
        \begin{tabular}{lccccccccc}
            \toprule
            \multirow{2}{*}{Method} & \multicolumn{3}{c}{Standard} & \multicolumn{3}{c}{Novel Category} & \multicolumn{3}{c}{Novel Camera} \\
            \cmidrule(lr){2-4} \cmidrule(lr){5-7} \cmidrule(lr){8-10}
                                                                        & DirAcc $\uparrow$ & FullAcc $\uparrow$ & DepthErr $\downarrow$ & DirAcc $\uparrow$ & FullAcc $\uparrow$ & DepthErr $\downarrow$ & DirAcc $\uparrow$ & FullAcc $\uparrow$ & DepthErr $\downarrow$ \\
            \midrule
            Qwen3-VL-8B~\cite{Qwen3-VL}                                 & 28.7              & 11.4               & 0.32                  & 27.5              & 10.9               & 0.37                  & 28.3              & 11.1               & 0.33 \\
            Qwen3-VL-8B~\cite{Qwen3-VL}$_{\text{\textit{w/}~Our Data}}$ & 75.6              & 55.4               & 0.21                  & 56.3              & 41.2               & 0.27                  & 68.7              & 49.8               & 0.23 \\
            \midrule
            \rowcolor{gray!15}
            \textbf{\methodName\ (Ours)}                                & \textbf{89.2}     & \textbf{76.7}      & \textbf{0.16}         & \textbf{74.4}     & \textbf{58.5}      & \textbf{0.19}         & \textbf{85.6}     & \textbf{73.2}      & \textbf{0.17} \\
            \bottomrule
        \end{tabular}
    }
\end{table}

\subsection{Improving Closed-Set 3D Detectors}
\label{sec:exp_closedset}
We first ask whether \ourMethod can improve the depth quality of strong, well-tuned closed-set detectors, even within their own training domain. We apply \ourMethod as a post-hoc refinement step on top of the recent state-of-the-art MonoCoP~\cite{zhang2025unleashing}, evaluated on the KITTI~\cite{geiger2012we} validation set under the standard \apThreeD protocol at IoU $\geq 0.7$.

\begin{wraptable}{r}{0.5\textwidth}
    \centering
    \vspace{-6mm}
    \small
    \caption{
        \textbf{Improving closed-set \threeD detectors} at IoU $\geq 0.7$. Applying \ourMethod on top of MonoCoP~\cite{zhang2025unleashing} yields consistent gains.
    }
    \resizebox{0.45\textwidth}{!}{
        \begin{tabular}{l|ccc}
            \toprule
            \multirow{2}{*}{Method} & \multicolumn{3}{c}{\textbf{\val}, \apThreeD ($\uparrow$)} \\
                                               & Easy           & Mod.           & Hard \\
            \midrule
            \monodetr~\cite{zhang2023monodetr} & 28.84          & 20.61          & 16.38 \\
            \monocd~\cite{yan2024monocd}       & 26.45          & 19.37          & 16.38 \\
            \fdThreeD~\cite{wu2024fd3d}        & 28.22          & 20.23          & 17.04 \\
            \monomae~\cite{jiang2024monomae}   & 30.29          & 20.90          & 17.61 \\
            \monotakd~\cite{liu2025monotakd}   & 34.36          & 22.61          & 19.88 \\
            \monodgp~\cite{pu2024monodgp}      & 30.76          & 22.34          & 19.02 \\
            MonoCoP~\cite{zhang2025unleashing} & 32.06          & 23.98          & 20.64 \\
            \rowcolor{gray!15}
            \textbf{\methodName\ (Ours)}       & \textbf{35.62} & \textbf{27.47} & \textbf{21.31} \\
            \bottomrule
        \end{tabular}
    }
    \label{tab:kitti_val}
\end{wraptable}

As shown in \cref{tab:kitti_val}, applying \ourMethod on top of MonoCoP yields consistent gains across all difficulty levels, raising \apThreeD from $32.06 \to 35.62$ on \textit{Easy}, $23.98 \to 27.47$ on \textit{Moderate}, and $20.64 \to 21.31$ on \textit{Hard}. The improvements are most pronounced on the Easy and Moderate splits, where less occluded objects allow the visual signature of depth misalignment to be read most reliably. The result confirms that even mature closed-set detectors leave object-level depth as an under-addressed source of error, and that \ourMethod recovers a substantial part of it without changing the underlying detector.

\begin{table}[t]
    \centering
    \small
    \caption{
        \textbf{Improving open-vocabulary \threeD detectors on \omniThreeD~\cite{brazil2023omni3d}.} \ourMethod refines DetAny3D~\cite{zhang2025detect} conditioned on ground-truth \twoD boxes, the strongest variant of DetAny3D, evaluated across the six \omniThreeD sub-datasets. \ourMethod lifts \apThreeD by $+4.35$ overall through depth refinement alone, demonstrating that the gain holds even on top of an oracle-grade detector.
    }
    \vspace{-2mm}
    \resizebox{1\linewidth}{!}{
        \begin{tabular}{l|ccccccc}
            \toprule
            Method                                                                     & ${\rm AP^{kit}_{3D}} \uparrow$ & ${\rm AP^{nus}_{3D}} \uparrow$ & ${\rm AP^{sun}_{3D}} \uparrow$ & ${\rm AP^{ark}_{3D}} \uparrow$ & ${\rm AP^{obj}_{3D}} \uparrow$ & ${\rm AP^{hyp}_{3D}} \uparrow$ & ${\rm AP_{3D}} \uparrow$ \\
            \midrule
            Cube R-CNN~\cite{brazil2023omni3d}                                         & 32.50                          & 30.06                          & 15.33                          & 41.73                          & 50.84                          & 7.48                           & 23.26 \\
            OVMono3D~\cite{yao2025open}                                                & 25.45                          & 24.33                          & 15.20                          & 41.60                          & 58.87                          & 7.75                           & 22.98 \\
            DetAny3D~\cite{zhang2025detect}                                            & 31.61                          & 30.97                          & 18.96                          & 46.13                          & 54.42                          & 7.17                           & 24.92 \\
            DetAny3D~\cite{zhang2025detect}$_{\text{\textit{w/}~Ground-Truth 2D Box}}$ & 38.68                          & 37.55                          & 46.14                          & 50.62                          & 56.82                          & 15.98                          & 34.38\\
            \midrule
            \rowcolor{gray!15}
            \textbf{\methodName (Ours)}                                                & \textbf{43.47}                 & \textbf{44.13}                 & \textbf{48.58}                 & \textbf{54.20}                 & \textbf{59.29}                 & \textbf{17.80}                 & \textbf{38.73} \\
            \bottomrule
        \end{tabular}
    }
    \label{tab:ov}
\end{table}

\subsection{Improving Open-Vocabulary 3D Detectors}
\label{sec:exp_openvocab}

\begin{figure}[t]
    \centering
    \includegraphics[width=\linewidth]{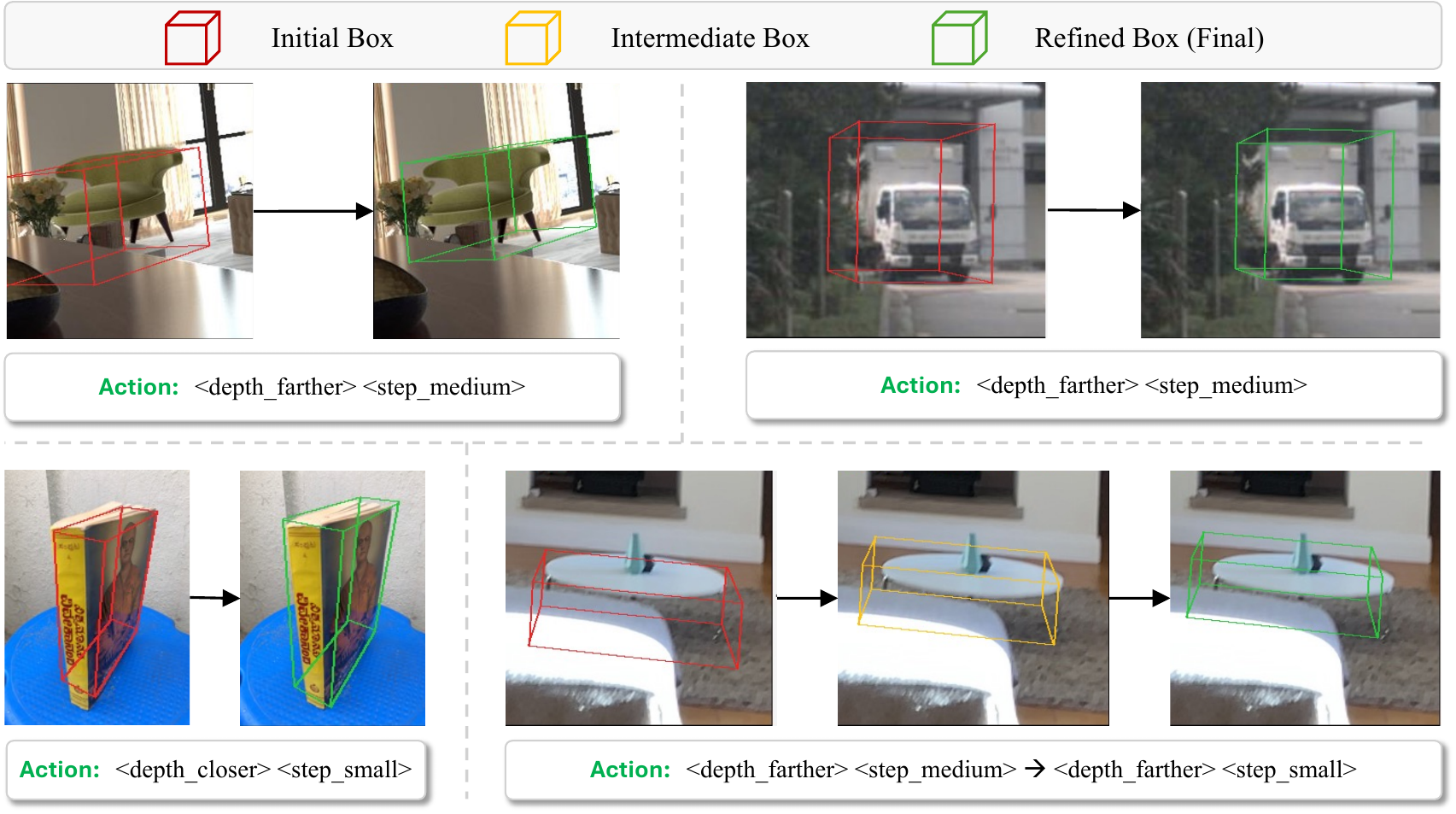}
    \small
    \caption{
        \textbf{Single-step and iterative depth refinement.}
        \ourMethod refines DetAny3D~\cite{zhang2025detect}'s initial prediction (red) toward the refined result (green). Most cases are corrected in a single step; in some cases the model applies multiple action steps until the projected box aligns with the target.
    }
    \label{fig:vis}
\end{figure}

Open-vocabulary detectors face a much harder regime: the upstream detector must localize categories it has never been trained on, and current methods compensate by relying on depth foundation models for \threeD geometry, the very source of bias we identified in \cref{sec:intro}. We test whether \ourMethod can correct this propagated bias by applying it on top of the state-of-the-art DetAny3D~\cite{zhang2025detect} across the full \omniThreeD~\cite{brazil2023omni3d} test split.
In \cref{tab:ov}, we apply \ourMethod on top of DetAny3D conditioned on ground-truth \twoD boxes ($34.38$ \apThreeD), the strongest variant of DetAny3D in which \twoD localization is already error-free. Even on this oracle baseline, refining only the object depth with \ourMethod lifts \apThreeD to $38.73$ ($+4.35$), with consistent gains on every \omniThreeD sub-dataset. The improvement comes entirely from depth correction, since all other box attributes are held fixed, demonstrating that \ourMethod's gains are not limited to weak detectors: it provides a meaningful and orthogonal correction even on top of an oracle-grade open-vocabulary detector.
\cref{fig:vis} illustrates this qualitatively, with most cases corrected in a single decisive step and a few  requiring a second pass; we quantify convergence behavior and inference cost in \cref{sec:appendix-inference}, and provide additional comparisons in \cref{sec:more-ov-results}.

\subsection{Improving 3D Auto-Labeling Methods}
\label{sec:exp_autolabel}

\begin{wraptable}{r}{0.45\linewidth}
    \vspace{-6mm}
    \centering
    \small
    \caption{
        \textbf{Improving \threeD auto-labeling methods.}
        \ourMethod refines LabelAny3D's pseudo-labels, narrowing the gap to GT-trained accuracy.
    }
    \label{tab:auto_label}
    \resizebox{\linewidth}{!}{
        \begin{tabular}{l|ccc}
            \toprule
            Method                              & Easy           & Mod.           & Hard \\
            \midrule
            GT~\cite{geiger2012we}              & 32.06          & 23.98          & 20.64 \\
            LabelAny3D~\cite{yao2025labelany3d} & 13.57          & 11.78          & 10.27 \\
            \rowcolor{gray!15}
            \textbf{\methodName\ (Ours)}        & \textbf{15.76} & \textbf{14.15} & \textbf{10.99} \\
            \bottomrule
        \end{tabular}
    }
    \vspace{-4mm}
\end{wraptable}
Beyond detection, depth foundation models also propagate their bias into auto-labeling pipelines that distill them into pseudo-\threeD annotations. To evaluate whether \ourMethod can correct this propagated bias, we adopt the following protocol: we use LabelAny3D~\cite{yao2025labelany3d} to auto-label a dataset with available GT, refine the resulting pseudo-labels with \ourMethod, and train a downstream \threeD detector~\cite{zhang2025unleashing} on the refined pseudo-labels. Detection accuracy on the held-out validation set, compared against a model trained on the original GT, measures how faithfully the auto-labeled data approximates the GT.

\cref{tab:auto_label} reports KITTI validation \apThreeD for detectors trained on LabelAny3D's pseudo-labels with and without \ourMethod refinement. Refining the pseudo-labels lifts the downstream detector by $+2.19$/$+2.37$/$+0.72$ on Easy / Moderate / Hard, narrowing the gap to GT-trained accuracy. Because the only change is a refinement pass on the auto-labels themselves, the gain isolates the contribution of \ourMethod's depth correction to label quality, and shows that refinement directly translates into better trained detectors downstream.

\subsection{Is a VLM Necessary?}
\label{sec:exp_nonvlm}

\begin{wraptable}{r}{0.5\linewidth}
    \vspace{-4mm}
    \centering
    \small
    \caption{
        \textbf{Non-VLM refinement mechanisms.}
        All methods refine MonoCoP's depth on \kitti val (\apThreeD, IoU $\geq 0.7$). Geometric fitting uses the GT \twoD box.
    }
    \label{tab:nonvlm}
    \resizebox{\linewidth}{!}{
        \begin{tabular}{l|ccc}
            \toprule
            Method                             & Easy           & Mod.           & Hard \\
            \midrule
            MonoCoP~\cite{zhang2025unleashing} & 32.06          & 23.98          & 20.64 \\
            Geometric fitting                  & 29.89          & 21.55          & 18.36 \\
            Geometric fitting (GT dims + yaw)  & 30.56          & 21.79          & 18.43 \\
            DINOv2 action classifier           & 24.71          & 17.62          & 12.78 \\
            \rowcolor{gray!15}
            \textbf{\methodName\ (Ours)}       & \textbf{35.62} & \textbf{27.47} & \textbf{21.31} \\
            \bottomrule
        \end{tabular}
    }
    \vspace{-4mm}
\end{wraptable}
We compare \ourMethod with two simpler mechanisms that could in principle judge image-space alignment, both refining only the depth of MonoCoP predictions (\cref{tab:nonvlm}). \emph{Geometric projection fitting} searches along the camera ray for the depth whose projected cuboid best overlaps the object's tight \twoD box, which we take from the ground truth to favor the baseline. It still lowers \apThreeD, even with GT dimensions and yaw: a tight \twoD box covers only the visible object, whereas the projected cuboid also spans occluded corners and empty volume, so the two objectives disagree; an $8\%$ extent mismatch already shifts a car at $40$\,m by about $3.3$\,m. A \emph{DINOv2 action classifier}~\cite{oquab2023dinov2}, trained on the same data to emit the same action tokens, degrades accuracy even more: its crop-to-class mapping does not transfer from pure depth perturbations to the coupled errors of a real detector, and every wrong token becomes a metric error. The gains of \ourMethod therefore come from the VLM's visual judgment rather than from the discrete action space itself; derivations and full details are in \cref{sec:appendix_baselines}.

\subsection{Ablation Studies}
\label{sec:ablation}

We ablate three design dimensions central to \ourMethod: how to surface the corrective action to the VLM, how to initialize the new action-token embeddings, and how to set up the training recipe. Each ablation in \cref{tab:ablation_main} compares our default against natural alternatives to support the design choices made in \cref{sec:action_tokens,sec:training_inference}. We report \textbf{DirAcc} (direction prediction accuracy) and \textbf{FullAcc} (both direction and magnitude correct).

\noindent\textbf{Action-token vocabulary.}
A natural alternative to introducing new tokens is to let the VLM verbalize the action in plain text and parse it back, or to introduce a single special token for each of the seven possible actions (six directional-magnitude combinations plus one standalone \texttt{ok} action). Plain text loses $13.6$ DirAcc and $21.3$ FullAcc compared to ours (\cref{tab:ablation_main}, left): free-form generation introduces unnecessary lexical variability, making the supervision target sparse and harder to align consistently across samples. Joint tokens, which dedicate one special token to each action, recover most of the gap on DirAcc ($86.8$) but still lag on FullAcc ($72.5$ vs.\ $76.7$): each joint token is supervised on only $\sim\!\tfrac{1}{7}$ of training samples, while our factored direction and magnitude tokens each see $\sim\!\tfrac{1}{3}$. Factoring direction and magnitude is therefore not just a clean abstraction; it provides substantially denser supervision for each token.

\noindent\textbf{Action-token initialization.}
The six new tokens carry no meaning at initialization, so how we seed them shapes what the embeddings can become through Stage~$1$ warm-up. We compare three options: \emph{random}; \emph{anchor}, which copies the embedding of ``depth'' into all six tokens; and \emph{semantic} (ours), where each token is initialized from a descriptive phrase (\cref{tab:action_tokens} in \cref{app:appendix_impl}). All three reach similar DirAcc after warm-up (within $2.4$ points), but on FullAcc random costs $3.5$ points and anchor costs $2.3$ relative to semantic (\cref{tab:ablation_main}, middle). The asymmetry reflects the structure of the two token groups. The three direction tokens are pairwise distinct categories, and random or identical initializations separate cleanly along this categorical axis during warm-up; DirAcc therefore largely recovers under any initialization. The three magnitude tokens, by contrast, form an \emph{ordinal} progression (small\,$\to$\,medium\,$\to$\,large); semantic initialization provides a better inductive bias for this ordinal structure from the start, whereas random or single-anchor initializations must recover it from scratch within the limited warm-up budget. FullAcc requires magnitude to be correct, which is precisely where this gap appears.

\noindent\textbf{Training recipe.}
Our training recipe makes two choices: it keeps the vision encoder frozen throughout, and it pairs each sample with a chain-of-thought (CoT) rationale rather than supervising on the action tokens alone. Both choices contribute substantially (\cref{tab:ablation_main}, right). Unfreezing the vision encoder costs $6.5$ DirAcc and $7.1$ FullAcc, likely because the encoder drifts away from the broad visual priors acquired during VLM pretraining and over-specializes to the narrow distribution of wireframe overlays. Removing the CoT rationale costs $4.7$ DirAcc and $6.5$ FullAcc: without CoT, the supervision becomes concentrated almost entirely on the action tokens themselves, losing the denser reasoning supervision that explicitly connects each action to visual evidence and making the model more susceptible to shortcut learning.

\begin{table}[t]
    \centering
    \small
    \caption{
        \textbf{Ablation studies across three design dimensions of \ourMethod.}
        Default configurations are highlighted in gray and outperform the alternatives.
    }
    \label{tab:ablation_main}
    \resizebox{\linewidth}{!}{%
        \begin{tabular}{c@{\hspace{6mm}}c@{\hspace{6mm}}c}
            \begin{tabular}{@{\hspace{2mm}}l@{\hspace{2mm}}c@{\hspace{2mm}}c@{\hspace{2mm}}}
                \toprule
                \textit{Action vocabulary}            & DirAcc                             & FullAcc \\
                \midrule
                Plain text                            & $75.6$                             & $55.4$ \\
                Joint tokens                          & $86.8$                             & $72.5$ \\
                \cellcolor{gray!12}\textbf{Dir + Mag} & \cellcolor{gray!12}$\mathbf{89.2}$ & \cellcolor{gray!12}$\mathbf{76.7}$ \\
                \bottomrule
            \end{tabular}
            &
            \begin{tabular}{@{\hspace{2mm}}l@{\hspace{2mm}}c@{\hspace{2mm}}c@{\hspace{2mm}}}
                \toprule
                \textit{Initialization}              & DirAcc                             & FullAcc \\
                \midrule
                Random                               & $86.8$                             & $73.2$ \\
                Anchor                               & $87.6$                             & $74.4$ \\
                \cellcolor{gray!12}\textbf{Semantic} & \cellcolor{gray!12}$\mathbf{89.2}$ & \cellcolor{gray!12}$\mathbf{76.7}$ \\
                \bottomrule
            \end{tabular}
            &
            \begin{tabular}{@{\hspace{2mm}}l@{\hspace{2mm}}c@{\hspace{2mm}}c@{\hspace{2mm}}}
                \toprule
                \textit{Training recipe}         & DirAcc                             & FullAcc \\
                \midrule
                No freeze vision                 & $82.7$                             & $69.6$ \\
                No CoT                           & $84.5$                             & $70.2$ \\
                \cellcolor{gray!12}\textbf{Ours} & \cellcolor{gray!12}$\mathbf{89.2}$ & \cellcolor{gray!12}$\mathbf{76.7}$ \\
                \bottomrule
            \end{tabular}
        \end{tabular}%
    }
    \vspace{-2mm}
\end{table}

\section{Conclusion}
\label{sec:conclusion}

We presented \ourMethod, which refines the depth of a monocular \threeD box after the fact by asking a vision-language model whether the projected box fits the object in the image. The model answers with a direction and a step size relative to the object's size, and repeating this moves the box until it fits, so no metric depth is ever regressed. As a post-hoc step, it improves MonoCoP on \kitti, DetAny3D with ground-truth \twoD boxes on \omniThreeD, and LabelAny3D pseudo-labels, whereas geometric fitting and a discriminative classifier given the same inputs make the boxes worse. More broadly, our results suggest that a model which cannot output a precise number can still reliably judge whether a rendered candidate is right. The approach does need the object to be visible enough to judge the fit, so heavily occluded or truncated objects remain hard, and its projection assumes a pinhole camera. Future work could combine the visual check with scene context or temporal cues across video frames, and extend it to omnidirectional cameras.

\clearpage

{
    \small
    \bibliographystyle{plain}
    \bibliography{reference}
}

\clearpage
\appendix
\crefalias{section}{appendix}

\section{Why Simpler Alternatives Fall Short}
\label{sec:appendix_baselines}

\ourMethod uses a VLM to judge image-space alignment. Two simpler mechanisms could in principle serve the same role: a purely geometric procedure that fits depth by projection, and a small discriminative network trained to emit the same action tokens. We implement both and find that neither recovers the improvement, for different reasons. All experiments in this section refine MonoCoP~\cite{zhang2025unleashing} predictions on \kitti~\cite{geiger2012we} validation under the \apThreeD protocol at IoU $\geq 0.7$ used in \cref{tab:kitti_val}, and modify only the object depth.

\begin{table}[t]
    \centering
    \small
    \caption{
        \textbf{Comparison with non-VLM refinement mechanisms.}
        All methods refine MonoCoP~\cite{zhang2025unleashing} predictions on \kitti validation (\apThreeD at IoU $\geq 0.7$) and modify only the object depth. Both geometric-fitting variants are given the \emph{ground-truth} tight \twoD box as the fitting target, an oracle input \ourMethod never receives.
    }
    \label{tab:appendix_baselines}
    \resizebox{\linewidth}{!}{
        \begin{tabular}{llcccc}
            \toprule
            Method                                         & Box attributes & Easy           & Mod.           & Hard           & $\Delta$ vs.\ MonoCoP \\
            \midrule
            MonoCoP~\cite{zhang2025unleashing} (unrefined) & predicted      & 32.06          & 23.98          & 20.64          & --- \\
            \midrule
            Geometric projection fitting                   & predicted      & 29.89          & 21.55          & 18.36          & $-2.17$/$-2.43$/$-2.28$ \\
            Geometric projection fitting                   & GT dims + yaw  & 30.56          & 21.79          & 18.43          & $-1.50$/$-2.19$/$-2.21$ \\
            DINOv2 ViT-S/14 action classifier              & predicted      & 24.71          & 17.62          & 12.78          & $-7.35$/$-6.36$/$-7.86$ \\
            \midrule
            \rowcolor{gray!15}
            \textbf{\ourMethod\ (Ours)}                    & predicted      & \textbf{35.62} & \textbf{27.47} & \textbf{21.31} & $+3.56$/$+3.49$/$+0.67$ \\
            \bottomrule
        \end{tabular}
    }
\end{table}

\subsection{Geometric Projection Fitting}
\label{sec:appendix_geo_fitting}

\textbf{Why the fitting objective is biased.}
Given predicted dimensions, yaw, and the camera ray, one may recover depth by searching for the value whose projected cuboid best overlaps the object's tight \twoD box. This resembles the alignment task we assign to the VLM, but the two objectives are not equivalent: a tight \twoD box encloses the \emph{visible} object pixels, whereas the projected cuboid also spans occluded corners and empty volume outside the visible silhouette. The mismatch is not a second-order effect. Consider a front-facing, on-axis cuboid with center depth $z^\star$, height $H$, and length $L$, under focal length $f$. Its projected envelope height at depth $z$ is $h_{\text{cub}}(z) = fH / (z - L/2)$, while the annotated tight box has height $h_{\text{2D}} = \rho\, h_{\text{cub}}(z^\star)$ for some extent ratio $\rho$ that is not $1$ in general. Matching heights yields a recovered depth $\hat{z}$ satisfying
\begin{equation}
    \hat{z} - z^\star = \Bigl(\tfrac{1}{\rho} - 1\Bigr)\Bigl(z^\star - \tfrac{L}{2}\Bigr),
    \label{eq:fitting_bias}
\end{equation}
so $\rho < 1$ pushes the recovered depth farther, $\rho > 1$ pulls it closer, and only $\rho = 1$ is exact---a condition that accurate \twoD localization does not imply. At detection scale this residual is decisive: an $8\%$ extent mismatch ($\rho = 0.92$) displaces a car centered at $40$\,m by roughly $3.3$\,m, far exceeding the longitudinal tolerance of the $\text{IoU}_{\text{3D}} \geq 0.7$ criterion at that range. A visible-object mask inherits the same mismatch, since it too covers only the visible extent. The same difficulty appears in prior \monoThreeD work: Deep3DBox~\cite{mousavian20173d} recovers translation from this projected-box constraint, and later methods~\cite{lu2021geometry, MonoRCNN_ICCV21} model the uncertainty of geometry-derived depth rather than treating it as exact.

\textbf{Setup.}
We implement the baseline as a one-dimensional search along the original predicted camera ray, selecting the depth that maximizes the \twoD IoU between the tight \twoD box and the axis-aligned envelope of the projected cuboid, with dimensions, yaw, and confidence held fixed. We use the \emph{ground-truth} tight \twoD box as the fitting target, an oracle input \ourMethod itself never receives, since a predicted \twoD box would only add localization noise to this target; the setting therefore favors the baseline. We report two variants: one using the detector's predicted dimensions and yaw, and one in which these are replaced by their GT values, isolating how much of the baseline's behavior is attributable to attribute noise. For the oracle variant, each prediction is matched to the same-class GT object with the highest \threeD IoU, falling back to the object with the nearest projected center when no same-class GT overlaps it. A sanity check confirms the implementation: fitting the GT projected envelope with GT attributes recovers the GT depth exactly.

\textbf{Results.}
Geometric fitting \emph{lowers} \apThreeD by $2.17 / 2.43 / 2.28$ on Easy / Moderate / Hard (\cref{tab:appendix_baselines}), while \ourMethod improves the same predictions by $3.56 / 3.49 / 0.67$. Supplying GT dimensions and yaw recovers only $0.67 / 0.24 / 0.07$ AP, leaving the baseline $1.50 / 2.19 / 2.21$ below the unrefined detector. Attribute noise is therefore not the primary cause: the degradation traces to the fitting objective itself, as \cref{eq:fitting_bias} predicts.

\subsection{A Lightweight Discriminative Action Classifier}
\label{sec:appendix_classifier}

\textbf{Setup.}
The second alternative keeps our action formulation but replaces the decision model. We train a classifier built on a pretrained DINOv2 ViT-S/14~\cite{oquab2023dinov2} encoder with two lightweight heads predicting the direction and magnitude tokens. It receives the same object crop with the projected wireframe that \ourMethod sees, and is trained on the same depth-perturbation data with the same target tokens. The vocabularies, object-relative step sizes, depth-update rule, and iterative procedure are identical, so the decision model is the only component that differs.

\textbf{Results.}
Despite the pretrained encoder and identical supervision, the classifier reduces \apThreeD by $7.35 / 6.36 / 7.86$ on Easy / Moderate / Hard (\cref{tab:appendix_baselines}), well below the unrefined detector, while \ourMethod improves the same predictions by $3.56 / 3.49 / 0.67$. The predefined discrete action space is therefore not by itself sufficient to reproduce the improvement: what distinguishes the two methods is the visual judgment used to select among the actions, not the actions themselves.

We attribute this to a distribution shift that the refinement loop amplifies. Training samples perturb depth on otherwise ground-truth boxes, so the classifier observes misalignments caused by depth alone; real detector outputs instead carry coupled residual errors in depth, dimensions, orientation, and projected position, which produce a considerably wider range of image--wireframe relationships. Because every predicted token pair is converted directly into a metric update, a decision model that does not transfer under this shift does more than fail to help: it moves initially reasonable predictions away from the target, consistent with the degradation we observe.

\subsection{Discussion: What the VLM Contributes}
\label{sec:appendix_baseline_discussion}

Neither alternative can reliably assess image--wireframe consistency under realistic detector residuals, but they fail through different mechanisms.

\textbf{Geometric fitting is limited by objective mismatch.} It substitutes a surrogate, rectangular overlap between a tight \twoD box and the projected cuboid envelope, for the judgment we actually want, namely whether the projected cuboid is consistent with the visible object. Maximizing that surrogate can favor a depth with better \twoD agreement but worse \threeD localization, and the failure persists under GT \twoD conditioning and oracle attributes.

\textbf{The classifier is limited by generalization.} It optimizes the right objective over the right action space, yet a task-specific discriminative mapping from crop to class does not survive the shift from controlled perturbations to the coupled residual errors of a real detector, and under this formulation every mispredicted token is spent as a metric error.

\textbf{\ourMethod pairs broad visual priors with a constrained correction interface.} The VLM is never asked to regress metric depth. It evaluates the relationship between the visible object and the projected cuboid (contour agreement, surface enclosure, internal projected structure, and the direction in which the mismatch points) rather than apparent size alone, and expresses the result through a small set of direction and magnitude tokens. The pretrained multimodal representation supports interpreting varied visual evidence, while the constrained vocabulary converts that judgment into bounded, object-relative metric updates. A further asymmetry is that \ourMethod can \emph{abstain}: the $\langle\texttt{depth\_ok}\rangle$ action leaves $37.8\%$ of objects unchanged (\cref{sec:appendix-inference}), whereas a fitting procedure always commits to its argmax. The improvement therefore does not arise from discretization alone, but from pairing the discrete interface with a decision model that generalizes the alignment judgment to realistic detector outputs.

\section{Implementation Details}
\label{app:appendix_impl}

\textbf{Model architecture.}
We build \ourMethod on top of Qwen3-VL-8B-Instruct~\cite{Qwen3-VL},
extending its tokenizer with six special action tokens that factor the
corrective action into three direction tokens and three magnitude
tokens (\cref{tab:action_tokens}). Each new embedding row is initialized
via \emph{semantic initialization}: we tokenize a short phrase
describing the token's meaning (right column of \cref{tab:action_tokens})
and mean-pool the corresponding sub-token embeddings of the frozen base
model, placing each new token in a region of the embedding space that
already encodes its intended concept. Magnitudes correspond to
object-relative shifts of $0.20\,s_{\text{obj}}$,
$0.55\,s_{\text{obj}}$, and $1.10\,s_{\text{obj}}$ for small, medium,
and large, where $s_{\text{obj}} = (w + h + l) / 3$ is the object's
mean linear extent.

\begin{table}[h]
    \centering
    \small
    \caption{Action-token vocabulary, inference contract, and semantic initialization. Each new token's embedding is initialized by mean-pooling the sub-token embeddings of its init phrase.}
    \label{tab:action_tokens}
    \begin{tabular}{@{}lll@{}}
        \toprule
        Token                     & Meaning at inference                                          & Init phrase \\
        \midrule
        \texttt{<depth\_closer>}  & move box towards camera (decrease center $z$)                 & \textit{``move closer to camera''} \\
        \texttt{<depth\_farther>} & move box away from camera (increase center $z$)               & \textit{``move farther from camera''} \\
        \texttt{<depth\_ok>}      & no correction needed                                          & \textit{``no depth correction needed''} \\
        \midrule
        \texttt{<step\_small>}    & apply $0.20 \times s$ m (midpoint of $[0.10, 0.30)$)          & \textit{``small step size''} \\
        \texttt{<step\_medium>}   & apply $0.55 \times s$ m (midpoint of $[0.30, 0.80)$)          & \textit{``medium step size''} \\
        \texttt{<step\_large>}    & apply $1.10 \times s$ m (representative for $[0.80, \infty)$) & \textit{``large step size''} \\
        \bottomrule
    \end{tabular}
\end{table}

\textbf{Training data and infrastructure.}
We train on the curated dataset of approximately $3.0$M (image, target-tokens) samples (\cref{sec:appendix_data_perturb,sec:appendix_data_cot}).  All training is performed on $8$ NVIDIA H100 ($80$~GB) GPUs in bfloat16 with DeepSpeed ZeRO-$3$ and FlashAttention-$2$, using AdamW with cosine learning-rate decay and standard auto-regressive cross-entropy loss applied to both the chain-of-thought tokens and the action tokens. Stage 1 takes approximately $1$ hour and Stage 2 takes approximately $90$ hours per configuration.

\subsection{Two-Stage Training Protocol}
\label{sec:appendix_training_protocol}

\textbf{Single-step supervision.}
Both training stages use single-step supervision. Each training sample consists of (i) an input image with the candidate \threeD box rendered as a wireframe overlay (\cref{sec:appendix_data_perturb}), (ii) a target chain-of-thought reasoning trace (\cref{sec:appendix_data_cot}), and (iii) the GT action tokens $(a_d, a_m)$ corresponding to the perturbation slot. We do not unroll the iterative update at training time. Each action is a local decision conditioned only on the current visual state $\mathbf{I}^{(t)}$ and is therefore independent of the trajectory leading up to it; teaching the model a mapping from any current visual state to its locally correct action is sufficient for multi-turn inference to chain such decisions correctly.

\textbf{Stage 1: action-token warm-up.}
The six action tokens are newly added to the VLM's vocabulary. Directly co-training them with the full VLM is unstable, since the loss must simultaneously shape both the new embeddings and the LLM's use of them, and the LLM cannot meaningfully attend to embeddings that are still essentially random. We therefore freeze every transformer block and train only the new token embedding rows (\texttt{embed\_tokens} and \texttt{lm\_head}) for one epoch on a stratified subsample, using AdamW with peak learning rate $5 \times 10^{-3}$, cosine schedule, and $5\%$ warmup. This stage is intentionally short and has no goal beyond stabilizing the new embeddings before joint optimization.

\textbf{Stage 2: joint fine-tuning.}
With the action-token embeddings warmed up, we fully fine-tune the language tower together with the action-token embeddings while keeping the vision encoder frozen. Freezing the vision encoder is a deliberate design choice: the encoder retains the general visual priors learned during VLM pretraining, which the trained \ourMethod relies on to generalize across novel categories, scenes, and cameras at test time. We empirically observed that unfreezing the vision encoder during Stage 2 hurts generalization on \benchmarkName, particularly on the novel-scene split. We train for one epoch over the full curated dataset using AdamW with peak learning rate $1 \times 10^{-5}$, cosine schedule, $3\%$ warmup, and gradient clipping at $1.0$.

\subsection{Multi-Turn Inference}
\label{sec:appendix_training_inference}

\textbf{Iterative refinement loop.}
At inference, \ourMethod is applied iteratively to refine an initial noisy box $\mathbf{b}^{(0)}$ produced by an upstream detector. At each step $t$, the loop performs four operations in sequence:
\begin{enumerate}
    \item Render the current box $\mathbf{b}^{(t)}$ as a wireframe on the input image to obtain $\mathbf{I}^{(t)} = \mathrm{render}(\mathbf{I},\ \pi(\mathbf{b}^{(t)};\mathbf{K}))$.
    \item Query the VLM with $\mathbf{I}^{(t)}$ and a fixed instruction prompt; the model first emits a chain-of-thought reasoning trace, then the action tokens $(a_d^{(t)}, a_m^{(t)})$.
    \item Compute the depth update $\Delta d^{(t)} = \mathrm{dir}(a_d^{(t)}) \cdot \mathrm{mag}(a_m^{(t)},\ s_{\text{obj}})$ and update the box: $d^{(t+1)} = d^{(t)} + \Delta d^{(t)}$, with $(\hat{x}^{(t+1)}, \hat{y}^{(t+1)})$ re-derived along the camera ray to keep the box's image location consistent with the new depth.
    \item Continue if $a_d^{(t)} \in \{\langle\texttt{depth\_closer}\rangle, \langle\texttt{depth\_farther}\rangle\}$; terminate if $a_d^{(t)} = \langle\texttt{depth\_ok}\rangle$ or if the maximum number of iterations $T_{\max}$ is reached. We use $T_{\max} = 2$ in all main experiments.
\end{enumerate}
The chain-of-thought reasoning trace is generated at every step but used only for action-token decoding; it is discarded after the action is parsed.

\textbf{Why train-inference asymmetry is consistent.}
\ourMethod is trained on single-step supervision but deployed in a multi-turn loop. This asymmetry is consistent because the action prediction at each step depends only on the current visual state $\mathbf{I}^{(t)}$, not on the trajectory of past actions. The conditional distribution $P(a^{(t)} \mid \mathbf{I}^{(t)})$ that the model learns at training is exactly the distribution it is queried with at each inference step, regardless of how many refinement iterations have already occurred. Single-step supervision therefore teaches a mapping that generalizes naturally to arbitrarily many inference steps, and multi-turn inference simply chains these per-step decisions until the model itself signals convergence via $\langle\texttt{depth\_ok}\rangle$.

\textbf{Decoding.}
We use greedy decoding for both the chain-of-thought reasoning and the action tokens, with the action-token positions restricted to the six special tokens via constrained decoding. This guarantees that each action prediction is one of the valid $(a_d, a_m)$ combinations and prevents the model from emitting natural-language tokens at action positions.

\textbf{Termination behavior.}
The over-representation of $\langle\texttt{depth\_ok}\rangle$ in the training distribution (\cref{sec:appendix_data_perturb}) gives the trained model a mild conservative bias: under uncertainty, it tends to terminate refinement rather than apply an aggressive correction. Empirically, refinement uses $1.62$ VLM calls per object on average under the safeguard cap $T_{\max} = 2$, and $15.8\%$ of objects reach the cap; these correspond to cases where the model oscillates between $\langle\texttt{depth\_closer}\rangle$ and $\langle\texttt{depth\_farther}\rangle$ near the alignment boundary.

\begin{table}[t]
    \centering
    \small
    \caption{Oracle study under the \omniThreeD evaluation protocol. Each row replaces one attribute of Cube R-CNN's predictions with either the ground truth or the output of a depth foundation model.}
    \label{tab:oracle_appendix}
    \begin{tabular}{lcc}
        \toprule
        \textbf{Variant}          & \textbf{\apThreeD} & \textbf{$\Delta$ vs.\ baseline} \\
        \midrule
        Baseline (Cube R-CNN)     & 36.00              & --- \\
        Oracle dimensions         & 35.91              & $-0.09$ \\
        Oracle angle              & 36.22              & $+0.22$ \\
        \midrule
        Foundation depth (MoGe-2) & 32.32              & $-3.68$ \\
        Oracle depth              & 65.92              & $+29.92$ \\
        \bottomrule
    \end{tabular}
\end{table}

\section{Oracle Study Details}
\label{sec:appendix_oracle}

We provide additional details for the oracle study reported in \cref{fig:teaser}. We use Cube R-CNN~\cite{brazil2023omni3d} trained on \omniThreeD as the baseline detector and report \apThreeD under the standard \omniThreeD evaluation protocol.

For each predicted box, we find its corresponding GT box via greedy assignment based on \threeD IoU, with a matching threshold of 0.5. Each GT box can match at most one prediction. Predictions that do not match any GT are kept unmodified, and unmatched GT boxes contribute to recall as usual. Confidence scores from the original detector are always preserved so that AP ranking is unaffected.

For each matched (prediction, GT) pair, we replace one or more attributes of the prediction with the corresponding GT values:
\begin{itemize}
    \item \textit{Oracle dimensions} replaces $(\hat{w}, \hat{h}, \hat{l})$ with $(w^*, h^*, l^*)$.
    \item \textit{Oracle angle} replaces $\hat{\theta}$ with $\theta^*$.
    \item \textit{Oracle depth} replaces the predicted object depth with the GT depth. Because changing depth alone while keeping the predicted $(\hat{x}, \hat{y})$ fixed would cause the box to drift in the image, we re-derive $(\hat{x}, \hat{y})$ from the camera ray of the original prediction and the new depth, so the projected box stays at the same image location and only depth changes.
    \item \textit{Foundation depth} replaces the predicted object depth with a depth derived from MoGe-2~\cite{wang2025moge}, a state-of-the-art depth foundation model. Since MoGe-2 predicts the depth of the visible object \emph{surface} rather than the object \emph{center} that \threeD detection requires, we apply a geometric correction rather than reading MoGe-2's output directly. Specifically, we (i) project the predicted \threeD center onto the image plane to locate its corresponding pixel, (ii) read MoGe-2's depth at that pixel as a surface-depth estimate, and (iii) add an offset---computed from the predicted object dimensions and orientation---that accounts for the distance, along the camera ray, between the surface entry point of the box and its center. This converts the per-pixel surface depth produced by MoGe-2 into a center depth comparable to what the detector regresses. As with oracle depth, $(\hat{x}, \hat{y})$ is then re-derived along the camera ray to keep the box's image location fixed.
\end{itemize}
All other attributes remain unchanged in each variant.

\cref{tab:oracle_appendix} reports the full numbers. The unmodified Cube R-CNN baseline reaches an \apThreeD of 36.00. Replacing the predicted dimensions or yaw angle with their ground-truth values changes AP by less than 0.3 points in either direction, indicating that these two attributes are already well predicted by the in-domain detector. Replacing the predicted depth with the ground-truth depth boosts AP to 65.92 (a $+29.92$ gain), nearly doubling the baseline. Replacing the predicted depth with the surface-to-center-corrected depth from MoGe-2, in contrast, reduces AP to 32.32 (a $3.68$ drop), worse than the baseline detector's own predictions.

Two observations emerge from these numbers. First, object depth is the single dominant bottleneck of monocular \threeD detection: among the three \threeD attributes we ablate, only depth shows a substantial gap to perfect prediction. The detector's predicted dimensions and angles are already accurate enough that perfecting them yields negligible gain, while perfecting depth alone nearly doubles the baseline AP. This motivates \ourMethod's design of refining only the depth and leaving other attributes untouched. Second, depth foundation models do not yet provide the precision needed to close this bottleneck. Although MoGe-2 demonstrates strong zero-shot performance on unseen scenes, its object-level depth estimates---even after the surface-to-center correction described above---remain less accurate than those of an in-domain detector. We attribute this to a mismatch in objective: depth foundation models are trained with dense, pixel-level supervision and are not optimized to be metrically precise at the object level. Under tight \threeD IoU thresholds, even small object-level biases translate into substantial AP drops.

\section{Data Curation Details}
\label{sec:appendix_data}

We construct our training set (\cref{fig:dataset-curation}) from public \threeD-annotated images in three stages: source dataset selection (\cref{sec:appendix_data_source}), per-annotation filtering (\cref{sec:appendix_data_filter}), and depth-perturbation sample synthesis (\cref{sec:appendix_data_perturb}). The first two stages yield $335{,}167$ filtered annotations with reliable \threeD boxes, visually unambiguous objects, and trustworthy category labels; the third stage synthetically perturbs each annotation along its camera ray to generate approximately $3.0$M (image, target-tokens) training samples that span all action-token outcomes.

\subsection{Source Datasets}
\label{sec:appendix_data_source}

We build on the \omniThreeD release~\cite{brazil2023omni3d}, which re-annotates six existing datasets into a shared COCO-style format under a common camera convention ($+x$ right, $+y$ down, $+z$ into the screen). The six sources are intentionally heterogeneous: KITTI and nuScenes provide outdoor driving scenes; Objectron contributes hand-held captures of common indoor objects; SUN RGB-D and ARKitScenes provide indoor scans; and Hypersim contributes photorealistic synthetic renderings. In total, this yields $3{,}303{,}211$ annotations across $234{,}152$ images, spanning $50$ unified categories ($96$ raw category names before the \omniThreeD taxonomy merge).

\subsection{Annotation Filtering}
\label{sec:appendix_data_filter}

The raw release contains many annotations unsuitable as positive training examples, including invalid \threeD boxes, behind-camera projections, tiny or heavily truncated objects, severe occlusion, mis-categorized labels, and rendering artifacts. We apply a two-stage filter: a geometric stage that uses metadata fields available in \omniThreeD, followed by a VLM-based semantic stage that catches errors no geometric rule can detect.

\textbf{Geometric and metadata filter.}
An annotation is retained only if it is marked as a valid \threeD box, projects in front of the camera, has a \twoD footprint of at least $32 \times 32$ pixels, and satisfies $\texttt{truncation} \le 0.05$ and $\texttt{visibility} \ge 0.8$ (relaxed to $\ge 0.6$ for nuScenes, whose visibility field is discretized into coarse bins). When both $\texttt{bbox2D\_tight}$ and $\texttt{bbox2D\_proj}$ are available, we additionally require their \twoD IoU to exceed $0.85$, and we reject annotations whose geometric truncation $1 - \mathrm{area}(\texttt{bbox2D\_trunc}) / \mathrm{area}(\texttt{bbox2D\_proj})$ exceeds $0.05$. After this stage, $466{,}032$ annotations across $162{,}002$ images remain ($14.1\%$ of the raw pool); per-dataset retention is reported in \cref{tab:dataset_filter}.

\textbf{VLM-based semantic filter.}
Geometric rules cannot detect mis-categorized labels, semantic occlusion by foreground clutter, blurry captures, or rendering artifacts. We add a semantic stage that queries a vision-language model on each surviving annotation. For each annotation we extract a $336 \times 336$ RGB crop centered on the object using $\texttt{bbox2D\_tight}$ (or $\texttt{bbox2D\_proj}$ when unavailable) expanded by a $1.5\times$ context margin, with the original \twoD box rendered in red as a visual anchor. We prompt Qwen3-VL-30B-A3B-Instruct~\cite{Qwen3-VL} to return a structured judgment of whether the object is recognizable, its predicted category among a $54$-class closed set, agreement with the claimed category, occlusion severity, image-edge truncation, bounding-box tightness, and overall image quality. We retain an annotation when the object is visible, the predicted and claimed categories are not in clear conflict, and none of occlusion, truncation, box tightness, or image quality is rated in its worst category. This stage retains $71.9\%$ of the geometrically filtered pool, leaving $335{,}167$ annotations.

\begin{table}[t]
    \centering
    \small
    \caption{Per-dataset retention through the two-stage filtering pipeline. The first ``\%'' column reports retention from raw to after the geometric filter; the second reports retention from after the geometric filter to after the VLM-based semantic filter.}
    \label{tab:dataset_filter}
    \setlength{\tabcolsep}{6pt}
    \begin{tabular}{lrrrrr}
        \toprule
        \textbf{Dataset} & \textbf{Anns (raw)}    & \textbf{After geom.} & \textbf{\%}   & \textbf{After VLM} & \textbf{\%} \\
        \midrule
        KITTI            & 50{,}514               & 10{,}932             & 21.6          & 10{,}023           & 91.7 \\
        nuScenes         & 356{,}438              & 150{,}240            & 42.2          & 90{,}440           & 60.2 \\
        Objectron        & 54{,}214               & 41{,}578             & 76.7          & 39{,}898           & 96.0 \\
        SUN RGB-D        & 78{,}514               & 31{,}444             & 40.0          & 23{,}278           & 74.0 \\
        ARKitScenes      & 476{,}501              & 94{,}642             & 19.9          & 53{,}677           & 56.7 \\
        Hypersim         & 2{,}287{,}030          & 137{,}196            & 6.0           & 117{,}851          & 85.9 \\
        \midrule
        \textbf{Total}   & \textbf{3{,}303{,}211} & \textbf{466{,}032}   & \textbf{14.1} & \textbf{335{,}167} & \textbf{71.9} \\
        \bottomrule
    \end{tabular}
\end{table}

\subsection{Depth-Perturbation Training Set}
\label{sec:appendix_data_perturb}

To turn the $335{,}167$ filtered annotations into supervision for \ourMethod, we synthetically perturb each GT box along the camera ray through its center to generate (image, target-tokens) training pairs that span all action-token outcomes.

\textbf{Per-annotation sampling schedule.}
For each filtered annotation we draw $9$ perturbation samples, one per slot of a sampling schedule (\cref{tab:perturb_schedule}) that exactly covers the action-token vocabulary. Each sample moves the box's center along this ray so that its depth changes by $\Delta z = r \cdot s_{\text{obj}}$, keeping the projected center fixed, where $s_{\text{obj}} = (w + h + l) / 3$ is the object's mean linear extent and $r$ is a slot-specific shift ratio sampled uniformly from a fixed range. Positive $\Delta z$ pushes the box farther from the camera, so the corrective action is $\langle\texttt{depth\_closer}\rangle$; negative $\Delta z$ pulls the box toward the camera, so the corrective action is $\langle\texttt{depth\_farther}\rangle$. Lateral position and orientation are kept at GT throughout, so every visible misalignment is purely a depth misalignment. The schedule allocates three slots to the $\langle\texttt{depth\_ok}\rangle$ outcome---an exact-zero shift and two narrow dead-zones around zero---reflecting that depth values inside this band are visually indistinguishable from the GT and should be reported as already aligned.

\begin{table}[t]
    \centering
    \small
    \caption{Perturbation sampling schedule. Each annotation produces $9$ samples, one per slot. The shift ratio $r$ is drawn uniformly from the listed range, and the actual center shift is $\Delta z = r \cdot s_{\text{obj}}$.}
    \label{tab:perturb_schedule}
    \setlength{\tabcolsep}{8pt}
    \begin{tabular}{cllc}
        \toprule
        \textbf{Slot} & \textbf{Direction}                      & \textbf{Magnitude}                    & \textbf{Shift ratio $r$} \\
        \midrule
        0             & $\langle\texttt{depth\_closer}\rangle$  & $\langle\texttt{step\_small}\rangle$  & $[+0.10,\ +0.30)$ \\
        1             & $\langle\texttt{depth\_closer}\rangle$  & $\langle\texttt{step\_medium}\rangle$ & $[+0.30,\ +0.80)$ \\
        2             & $\langle\texttt{depth\_closer}\rangle$  & $\langle\texttt{step\_large}\rangle$  & $[+0.80,\ +1.50]$ \\
        3             & $\langle\texttt{depth\_farther}\rangle$ & $\langle\texttt{step\_small}\rangle$  & $[-0.30,\ -0.10)$ \\
        4             & $\langle\texttt{depth\_farther}\rangle$ & $\langle\texttt{step\_medium}\rangle$ & $[-0.80,\ -0.30)$ \\
        5             & $\langle\texttt{depth\_farther}\rangle$ & $\langle\texttt{step\_large}\rangle$  & $[-1.50,\ -0.80]$ \\
        6             & $\langle\texttt{depth\_ok}\rangle$      & ---                                   & $0$ \\
        7             & $\langle\texttt{depth\_ok}\rangle$      & dead-zone $+$                         & $[0,\ +0.10)$ \\
        8             & $\langle\texttt{depth\_ok}\rangle$      & dead-zone $-$                         & $(-0.10,\ 0]$ \\
        \bottomrule
    \end{tabular}
\end{table}

\textbf{Inference contract for magnitude tokens.}
At inference, each magnitude token corresponds to a fixed corrective shift used by the iterative refinement step. We use the bucket midpoint of each slot's shift-ratio range, so that $\langle\texttt{step\_small}\rangle$ applies $0.20 \cdot s_{\text{obj}}$, $\langle\texttt{step\_medium}\rangle$ applies $0.55 \cdot s_{\text{obj}}$, and $\langle\texttt{step\_large}\rangle$ applies $1.10 \cdot s_{\text{obj}}$ as a representative value for the open-tail $[0.80, \infty)$ regime.

\textbf{Image rendering.}
Each of the nine perturbation samples reuses the same $336 \times 336$ crop computed in \cref{sec:appendix_data_filter} from the unperturbed \twoD box. On top of this fixed crop we render a wireframe of the \emph{perturbed} cuboid: the eight corners of the GT box are translated with its center along the camera ray, projected through the original intrinsics $\mathbf{K}$, mapped into crop pixels, and connected by the twelve cuboid edges in green. Holding the crop region fixed across all nine slots is intentional: the only visual difference between samples of the same object is the wireframe itself, which prevents the model from exploiting crop-region cues as a shortcut for inferring the depth direction.

\textbf{Geometric rejection.}
A perturbation is rejected and re-sampled if any cuboid corner falls behind the camera or if the projected cuboid becomes degenerately small in the crop, guarding against samples in which the perturbed box is invisible or unreasonably tiny.

\textbf{Final perturbation pool.}
The full pipeline produces approximately $3.0$M (image, target-tokens) training samples from the $335{,}167$ distinct objects. By construction, the action-token distribution is uniform over the seven distinct token combinations: each of the six $\langle\texttt{depth\_closer}\rangle / \langle\texttt{depth\_farther}\rangle$ direction-magnitude pairs appears in $\frac{1}{9}$ of the samples, and $\langle\texttt{depth\_ok}\rangle$ accounts for the remaining $\frac{3}{9} \approx 33\%$ (\cref{tab:perturb_distribution}). The deliberate over-representation of $\langle\texttt{depth\_ok}\rangle$ introduces a mild conservative bias at inference: under uncertainty, the model prefers to terminate refinement, which costs at most one extra iteration while preventing unnecessary perturbation of already-aligned boxes.

\begin{table}[t]
    \centering
    \small
    \caption{Action-token distribution in the perturbation training set. The six direction--magnitude combinations are matched in count, and $\langle\texttt{depth\_ok}\rangle$ is over-represented to encourage conservative termination at inference.}
    \label{tab:perturb_distribution}
    \setlength{\tabcolsep}{8pt}
    \begin{tabular}{lrr}
        \toprule
        \textbf{Token combination}                                                   & \textbf{\# samples} & \textbf{Share} \\
        \midrule
        $\langle\texttt{depth\_closer}\rangle\,\langle\texttt{step\_small}\rangle$   & $\approx 335$K      & $11.1\%$ \\
        $\langle\texttt{depth\_closer}\rangle\,\langle\texttt{step\_medium}\rangle$  & $\approx 335$K      & $11.1\%$ \\
        $\langle\texttt{depth\_closer}\rangle\,\langle\texttt{step\_large}\rangle$   & $\approx 335$K      & $11.1\%$ \\
        $\langle\texttt{depth\_farther}\rangle\,\langle\texttt{step\_small}\rangle$  & $\approx 335$K      & $11.1\%$ \\
        $\langle\texttt{depth\_farther}\rangle\,\langle\texttt{step\_medium}\rangle$ & $\approx 335$K      & $11.1\%$ \\
        $\langle\texttt{depth\_farther}\rangle\,\langle\texttt{step\_large}\rangle$  & $\approx 335$K      & $11.1\%$ \\
        $\langle\texttt{depth\_ok}\rangle$                                           & $\approx 1.0$M      & $33.3\%$ \\
        \midrule
        Total                                                                        & $\approx 3.0$M      & $100.0\%$ \\
        \bottomrule
    \end{tabular}
\end{table}

\begin{figure}[t]
    \centering
    \includegraphics[width=0.98\linewidth]{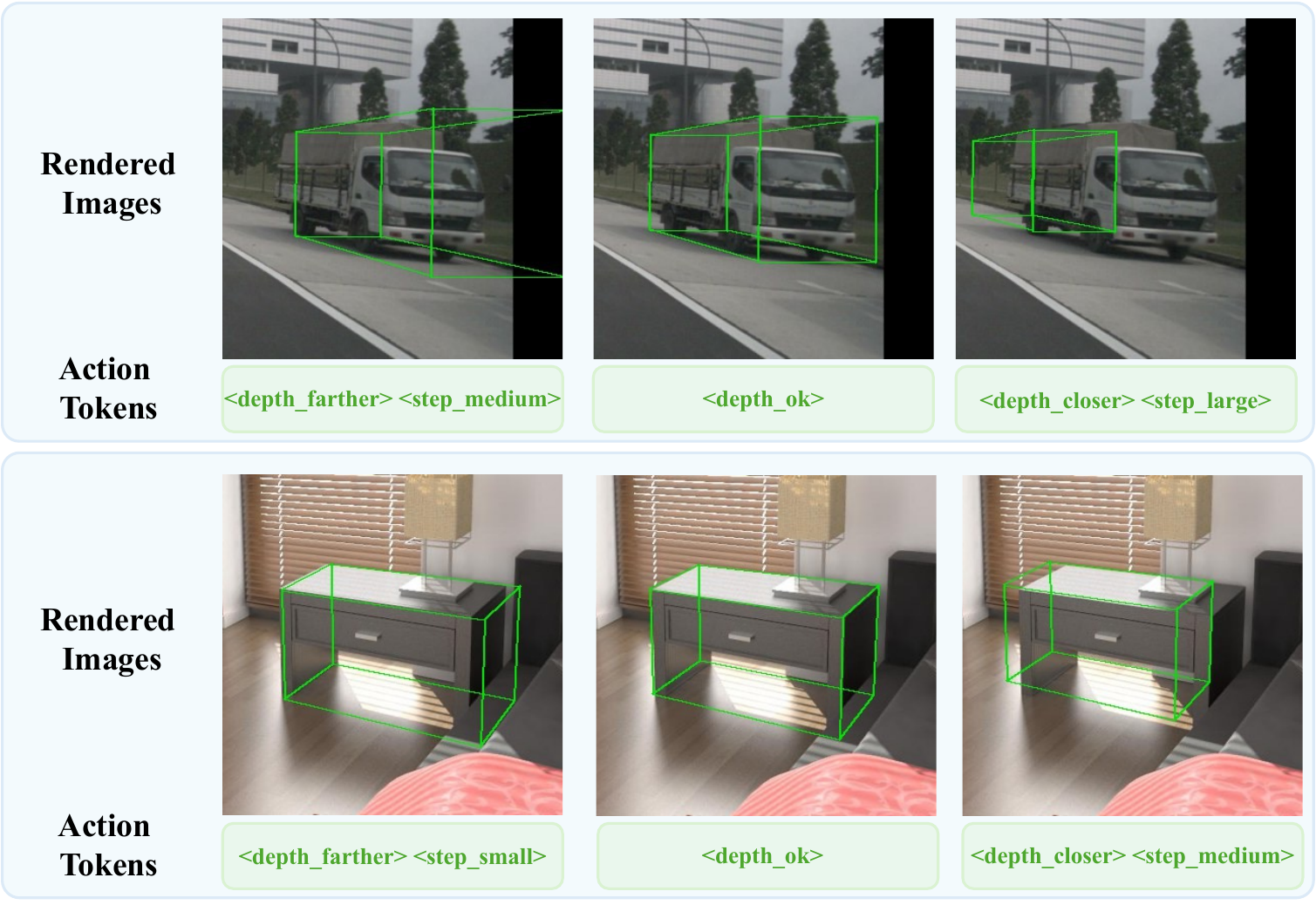}
    \small
    \caption{
        \textbf{Examples of generated perturbed data.}
    }
    \label{fig:perturb_examples}
\end{figure}

\textbf{Visualization.}
\cref{fig:perturb_examples} shows three sample crops drawn from a single object: slot 6 ($\langle\texttt{depth\_ok}\rangle$), slot 1 ($\langle\texttt{depth\_closer}\rangle\,\langle\texttt{step\_medium}\rangle$), and slot 4 ($\langle\texttt{depth\_farther}\rangle\,\langle\texttt{step\_medium}\rangle$). The three samples share the same crop region but differ only in the rendered wireframe, illustrating how a depth perturbation manifests visually as a wireframe that aligns with the object, sits in front of it, or sits behind it.

\subsection{Chain-of-Thought Reasoning Targets}
\label{sec:appendix_data_cot}

In preliminary experiments, fine-tuning the VLM to emit only action tokens led to a noticeable drop in its general instruction-following ability. We therefore pair every training sample with a short chain-of-thought (CoT) prefix that the model produces before emitting the action tokens. Keeping the CoT short is also consistent with recent evidence that longer textual reasoning lowers MLLM accuracy on fine-grained visual classification~\cite{zhu2026can}. The CoT acts as a language-anchored intermediate signal: the model articulates which object it is looking at, what visual cue it is reading, and which direction the cue implies, before committing to a discrete action.

\textbf{Generator choice.}
We synthesize each CoT with a two-model pipeline that splits the work by what each model can reliably do.
A VLM (Qwen3-VL-235B-A22B-Instruct~\cite{Qwen3-VL}) handles \emph{object identification and description}: given the image crop, it produces a short caption naming the target object and noting salient appearance cues (\eg, ``a black car viewed from behind on a street''), a task on which zero-shot VLMs are reliable.
A text-only LLM (Qwen3-235B-A22B-Instruct-2507~\cite{qwen3}) then takes this caption together with the sample metadata (object category, GT direction, GT magnitude token) and composes the full CoT, including the optional world-knowledge size recall, the visual-evidence framing of the misalignment, and the corrective decision.
We deliberately keep the alignment judgment itself out of the VLM at this stage: reliably telling a well-fitted \threeD wireframe from a poorly-fitted one is precisely the capability we aim to teach \ourMethod, so soliciting that judgment from a zero-shot VLM would inject unreliable supervision into training.

\textbf{Reasoning structure.}
Each CoT follows a fixed four-step structure that mirrors how a human annotator would reason about a perturbed wireframe.

\begin{itemize}
    \item \textit{Identify.} The model first names the object visible in the crop, optionally with a brief scene hint (\eg, ``A chair viewed slightly from above.'').
    \item \textit{Recall} (optional, skipped in $30\%$ of samples). The model recalls a generic, world-knowledge size range for the object category in approximate units (\eg, ``Chairs typically span around 60\,cm wide and 90--110\,cm tall.''). Precise metric values are forbidden, and the recalled range is decoupled from the actual GT.
    \item \textit{Visual evidence.} The model describes the misalignment between wireframe and object using either the size cue (\eg, ``The wireframe is noticeably smaller than the chair's outline.''), the enclosure cue (\eg, ``The chair's near surface protrudes in front of the wireframe's near face.''), or both, depending on the framing label drawn for this sample. Specific camera-relative distances are forbidden.
    \item \textit{Decision.} The model concludes by stating the implied depth error and the corrective action, consistent with the GT direction and magnitude token (\eg, ``This indicates the box is too far from the camera, requiring a closer correction by a medium step.'').
\end{itemize}

\textbf{Visual cue diversification.}
Depth error is observable through two complementary visual cues, and we deliberately diversify the CoTs across both so that the trained VLM learns each independently:
\begin{itemize}
    \item \textit{Size cue.} A box that is too far appears smaller than the object; too close, larger.
    \item \textit{Enclosure cue.} A box that is too far fails to enclose the object, with the object's near surface protruding past the wireframe; too close, the wireframe encloses the object loosely, leaving empty volume.
\end{itemize}
For each sample we draw a framing label from $\{\textit{size-only},\ \textit{enclosure-only},\ \textit{mixed}\}$ with weights $0.4 / 0.4 / 0.2$, and the prompt for non-mixed framings forbids the opposite framing's vocabulary, ensuring each CoT exercises a clean single-cue signal.

\textbf{Training-time use.}
At training time, each sample's target string is the concatenation of the generated reasoning prose and the canonical \texttt{Action} line, and the cross-entropy loss is applied uniformly over both. At inference, the reasoning prose is generated first and discarded; only the parsed \texttt{Action} tokens drive the iterative refinement step.

\section{\benchmarkName Benchmark Details}
\label{app:appendix_benchmark}

\benchmarkName is a controlled benchmark we construct on top of \omniThreeD~\cite{brazil2023omni3d} to evaluate \threeD box refinement as a standalone capability. Each test sample consists of a ground-truth \threeD box paired with a synthetically perturbed noisy version, and the task is to refine the noisy box back toward the GT.
The benchmark spans two distribution-shift axes plus an in-distribution reference, which jointly probe the generalization properties \ourMethod aims to achieve. We describe the construction of each split and the unified evaluation protocol below.

\subsection{Test-Sample Construction}
\label{sec:appendix_benchmark_samples}

For each \threeD annotation in the underlying source data, we generate noisy boxes following the same depth-perturbation protocol as the training set (\cref{sec:appendix_data_perturb}): the GT box's center is shifted along the camera ray by $\Delta z = r \cdot s_{\text{obj}}$, with $r$ drawn uniformly from each of the perturbation slots in the schedule. All other attributes (lateral position, dimensions, orientation) are held at GT, so each test sample isolates a depth misalignment of known direction and magnitude. Test samples are stratified across the perturbation schedule so that each direction-magnitude combination is equally represented, preventing any setting from being dominated by samples of a particular noise pattern.

\subsection{Standard}
\label{sec:appendix_benchmark_standard}

The Standard split serves as the in-distribution reference. Both training and test pools use the same \omniThreeD category vocabulary and camera intrinsics distribution. Training samples are drawn from \omniThreeD's standard train split; test samples are drawn from \omniThreeD's standard test split, following the construction protocol in \cref{sec:appendix_benchmark_samples}.

\subsection{Novel Category}
\label{sec:appendix_benchmark_category}

We follow the open-vocabulary monocular \threeD detection setup of \ovMonoThreeD~\cite{yao2025open}, which partitions the \omniThreeD category vocabulary into a base set and a novel set. During training, the data curation pipeline (\cref{sec:appendix_data}) drops every annotation belonging to a novel category, so the trained \ourMethod has never been supervised on these classes. At test time, we draw test samples (per \cref{sec:appendix_benchmark_samples}) from \omniThreeD's standard test split, restricted to the held-out novel categories.

\subsection{Novel Camera}
\label{sec:appendix_benchmark_camera}

To probe robustness to unseen camera intrinsics without changing the scene content, we synthetically rescale the original \omniThreeD test images while updating the camera intrinsics consistently. For each test image $\mathbf{I}$ with intrinsics $\mathbf{K}$, we apply an isotropic scale factor $s$ to obtain a rescaled image $\mathbf{I}_s$ and the rescaled intrinsics
\[
\mathbf{K}_s = \mathrm{diag}(s, s, 1)\,\mathbf{K},
\]
which preserves the geometric validity of the \threeD-to-\twoD projection but shifts the distribution of projected box sizes away from the training distribution.

Image rescaling alters the effective focal length while keeping the underlying scene unchanged: from the model's perspective, the same physical object now appears at a different projected size on the image plane. Because all other factors (object identity, scene composition, occlusion) are held constant, any drop in performance can be attributed unambiguously to camera-induced distribution shift. Training uses the unscaled \omniThreeD train split; test samples are drawn from the rescaled \omniThreeD test split.

\subsection{Evaluation Metrics}
\label{sec:appendix_benchmark_metrics}

We report three metrics that jointly characterize refinement quality.

\textbf{Direction accuracy (DirAcc).}
The fraction of test samples for which the refiner predicts the correct corrective direction token $a_d \in \{\langle\texttt{depth\_closer}\rangle,\langle\texttt{depth\_ok}\rangle,\langle\texttt{depth\_farther}\rangle\}$ at the first refinement step. This metric isolates the qualitative correctness of the refiner's first-step decision from the magnitude prediction, providing a clean proxy for the model's underlying alignment reasoning. Higher is better.

\textbf{Full action accuracy (FullAcc).}
The fraction of test samples for which the refiner predicts both the correct direction $a_d$ and the correct magnitude $a_m$ at the first refinement step. FullAcc strictly extends DirAcc by additionally requiring the magnitude bucket (small, medium, or large) to match the ground-truth bucket induced by the perturbation. Higher is better.

\textbf{Depth error (DepthErr).}
The mean absolute residual between the refined and GT object depths after the iterative refinement loop terminates:
\[
\mathrm{DepthErr} = \frac{1}{N} \sum_{i=1}^{N} \left| d_i^{(\text{refined})} - d_i^{*} \right|,
\]
where $d_i^{*}$ is the GT depth and $d_i^{(\text{refined})}$ is the depth after refinement. To prevent objects at different scales from dominating the average, the per-sample residual is normalized by $s_{\text{obj}}$ before averaging. Lower is better.

\textbf{Why three metrics together.}
DirAcc and FullAcc characterize the per-step prediction quality at the token level, while DepthErr captures the end-task refinement outcome after the iterative loop. A method can score high on DirAcc yet still leave residual depth error through biased magnitudes; conversely, a method can produce low DepthErr by chance through cancellation of errors across iterations without making genuinely correct per-step decisions. Reporting all three lets readers distinguish whether a method is failing to choose the direction, to choose the step size, or to converge to the right answer.

\section{Inference Analysis}
\label{sec:appendix-inference}

\ourMethod is trained with single-step supervision but applied iteratively at inference (\cref{sec:training_inference}): at each step, the model emits an action token that updates the candidate box's depth, until either the model commits to $\langle\texttt{depth\_ok}\rangle$ or a safeguard cap of $T_{\max}$ VLM calls is reached. We use $T_{\max}=2$ in all main experiments. Two natural questions arise: how quickly does the loop converge in practice, and what is the actual inference cost in a deployment setting? We answer both below.

\subsection{Convergence Behavior}

We sweep the iteration count $K \in \{0, 1, 2, 3, 4\}$ on KITTI~\cite{geiger2012we} validation Moderate, refining MonoCoP~\cite{zhang2025unleashing}'s predictions as the upstream detector. $K=0$ corresponds to the unrefined MonoCoP baseline, and $K \geq 1$ forces the loop to run exactly $K$ steps regardless of self-termination.

\begin{figure}[t]
    \centering
    \includegraphics[width=0.55\linewidth]{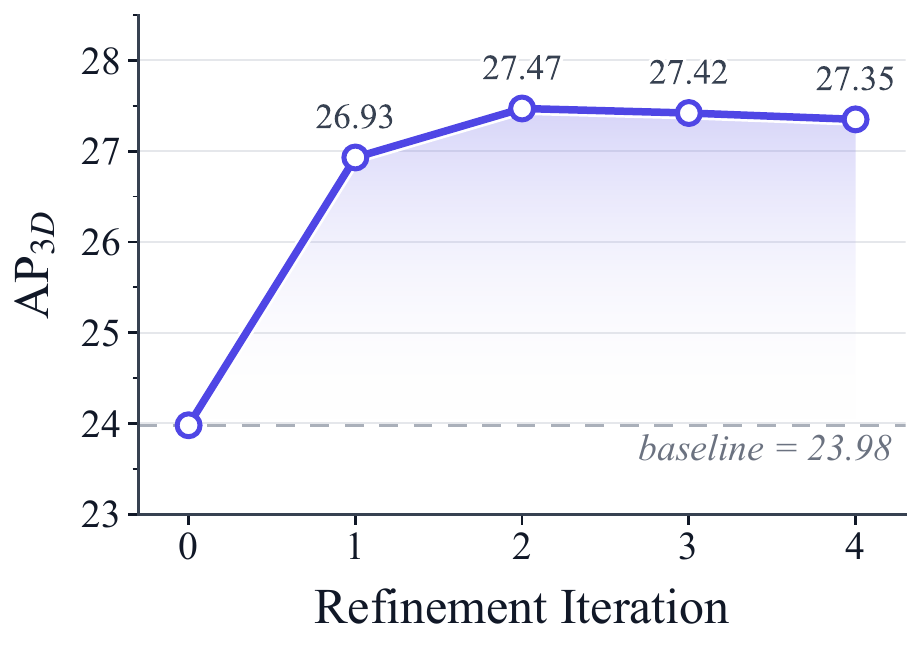}
    \small
    \caption{
        \textbf{Iterative refinement convergence on KITTI Moderate.}
        \apThreeD as a function of forced iteration count $K$. A single refinement step recovers most of the gain ($+2.95$ \apThreeD over the baseline); the second step adds another $+0.54$, after which performance plateaus and slightly drifts down.
    }
    \label{fig:iter_kitti}
\end{figure}

\cref{fig:iter_kitti} shows three regimes. \textit{First step} ($K=0 \to 1$): \apThreeD jumps from $23.98$ to $26.93$, a $+2.95$ gain that accounts for $\sim\!85\%$ of the total improvement. \textit{Second step} ($K=1 \to 2$): an additional $+0.54$ \apThreeD ($26.93 \to 27.47$), addressing the subset of objects whose alignment cannot be diagnosed from a single look at the wireframe. \textit{Beyond two steps} ($K=2 \to 4$): \apThreeD plateaus and drifts down slightly ($27.47 \to 27.35$), as a small fraction of already-aligned boxes is over-corrected. Based on this trade-off we set $T_{\max}=2$, which captures the peak accuracy at the lowest inference cost.

\subsection{Iteration-Count Distribution under Self-Termination}

With self-termination via $\langle\texttt{depth\_ok}\rangle$ enabled, each object is processed within $T_{\max}=2$ VLM calls. On KITTI validation, $37.8\%$ of objects exit on the first call (the model judges the upstream prediction already aligned and applies no update), $46.4\%$ exit on the second call after one applied update, and the remaining $15.8\%$ apply two updates and reach the cap. The mean number of VLM calls per object is $1.62$ under $T_{\max} = 2$, indicating that the safeguard cap is exercised only on the minority of harder cases.

\subsection{Inference Latency}

We measure absolute latency on a single NVIDIA H100 ($80$~GB) GPU with bfloat16 precision and FlashAttention-2. The wireframe rendering step is negligible ($<\!5$\,ms on CPU), and one VLM forward over the rendered object crop takes $\sim\!320$\,ms on average. Combined with the empirical mean of $1.62$ VLM calls per object, the average per-object refinement latency is $\sim\!520$ms. At the typical KITTI validation density of $\sim\!5.4$ objects per image (measured on the validation split), this corresponds to an end-to-end refinement overhead of $\sim\!2.8$s per image.

This overhead is non-trivial relative to the upstream detector itself but acceptable in offline use cases such as auto-labeling, dataset curation, and post-hoc accuracy improvement, where throughput is not the bottleneck. Two factors keep the cost bounded. First, refinement operates on localized object crops rather than the full image, so cost scales with the number of high-confidence proposals rather than image resolution. Second, the $\langle\texttt{depth\_ok}\rangle$ self-termination signal lets the model skip refinement entirely on $\sim\!38\%$ of objects, keeping the empirical mean below $T_{\max}$.

\section{More Results on Improving Open-Vocabulary 3D Detectors}
\label{sec:more-ov-results}

\cref{sec:exp_openvocab} reports our main open-vocabulary result on top of DetAny3D~\cite{zhang2025detect} conditioned on ground-truth \twoD boxes, an oracle setup chosen to isolate the contribution of depth refinement. To complement that stress-test, we also evaluate \ourMethod under a more realistic configuration in which DetAny3D is conditioned on \twoD boxes predicted by Cube R-CNN~\cite{brazil2023omni3d}, the standard closed-set detector used in the \omniThreeD benchmark. This setup reflects how an open-vocabulary refinement pipeline would be deployed in practice: an upstream \twoD detector proposes object regions, a 3D detector lifts them to \threeD, and \ourMethod refines the resulting depth.

\cref{tab-more-ov} reports \apThreeD on each of the six \omniThreeD sub-datasets. Applying \ourMethod on top of DetAny3D with Cube R-CNN's predicted \twoD boxes lifts overall \apThreeD from $24.92$ to $26.96$ ($+2.04$), with consistent improvements across every sub-dataset. The gain is smaller in absolute terms than in the oracle-2D setting ($+4.35$) because the upstream Cube R-CNN \twoD detections themselves carry localization error that propagates into the \threeD prediction and limits the headroom available to depth refinement. Even so, the consistent positive gains under noisy \twoD conditioning indicate that \ourMethod's refinement signal does not require clean upstream input: it transfers from the oracle setting to a fully realistic pipeline without retraining.

\begin{table}[tbp]
    \centering
    \vspace{-2mm}
    \caption{
        \textbf{Improving open-vocabulary \threeD detectors on \omniThreeD~\cite{brazil2023omni3d} under a realistic \twoD-box source.}
        \ourMethod refines DetAny3D~\cite{zhang2025detect} conditioned on Cube R-CNN's~\cite{brazil2023omni3d} predicted \twoD boxes, complementing the oracle-\twoD evaluation in \cref{sec:exp_openvocab}.
    }
    \vspace{-2mm}
    \resizebox{1\linewidth}{!}{
        \begin{tabular}{l|ccccccc}
            \toprule
            Method                             & ${\rm AP^{kit}_{3D}} \uparrow$ & ${\rm AP^{nus}_{3D}} \uparrow$ & ${\rm AP^{sun}_{3D}} \uparrow$ & ${\rm AP^{ark}_{3D}} \uparrow$ & ${\rm AP^{obj}_{3D}} \uparrow$ & ${\rm AP^{hyp}_{3D}} \uparrow$ & ${\rm AP_{3D}} \uparrow$ \\
            \midrule
            Cube R-CNN~\cite{brazil2023omni3d} & 32.50                          & 30.06                          & 15.33                          & 41.73                          & 50.84                          & 7.48                           & 23.26 \\
            OVMono3D~\cite{yao2025open}        & 25.45                          & 24.33                          & 15.20                          & 41.60                          & 58.87                          & 7.75                           & 22.98 \\
            DetAny3D~\cite{zhang2025detect}    & 31.61                          & 30.97                          & 18.96                          & 46.13                          & 54.42                          & 7.17                           & 24.92 \\
            \midrule
            \rowcolor{gray!15}
            \textbf{\methodName (Ours)}        & \textbf{33.94}                 & \textbf{32.29}                 & \textbf{21.07}                 & \textbf{47.83}                 & \textbf{56.36}                 & \textbf{9.05}                  & \textbf{26.96} \\
            \bottomrule
        \end{tabular}
    }
    \label{tab-more-ov}
\end{table}

\section{Preserving General Visual Capability}
\label{sec:exp_cvbench}

A practical concern is that fine-tuning may erode the VLM's general visual competence. Fine-tuning is known to affect a model beyond its target task: supervised safety fine-tuning of VLMs can reinforce spurious correlations between surface text patterns and the intended responses~\cite{chen2025safety}, and weight updates such as machine unlearning leave traces that are detectable from model outputs even on unrelated inputs~\cite{chen2025unlearning}. On CV-Bench~\cite{tong2024cambrian1} (\cref{tab:cvbench}), \ourMethod retains nearly all of the underlying Qwen3-VL-8B's capability: the overall score changes by only $1.41$ points ($85.94 \to 84.53$), with no sub-task degrading by more than $3$ points. This corroborates freezing the vision encoder throughout training (\cref{sec:training_inference}), which preserves the base model's broad visual priors while adding refinement as a new capability on top.

\begin{table}[t]
    \centering
    \caption{
        \textbf{General visual capability on CV-Bench~\cite{tong2024cambrian1}.}
        \ourMethod retains nearly all of the underlying Qwen3-VL-8B's capability, indicating that fine-tuning for refinement does not erode general visual reasoning.
    }
    \label{tab:cvbench}
    \vspace{-2mm}
    \resizebox{0.8\linewidth}{!}{
        \begin{tabular}{lccccc}
            \toprule
            Method                          & Depth & Distance & Relation & Count & Overall \\
            \midrule
            Qwen2.5-VL-7B~\cite{Qwen2.5-VL} & 86.33 & 75.83    & 88.92    & 64.85 & 78.17 \\
            Qwen3-VL-8B~\cite{Qwen3-VL}     & 94.17 & 87.00    & 93.54    & 72.59 & 85.94 \\
            \rowcolor{gray!15}
            \textbf{\methodName\ (Ours)}    & 93.83 & 84.00    & 92.77    & 71.07 & 84.53 \\
            \bottomrule
        \end{tabular}
    }
\end{table}

%
%

\section{Limitations and Future Work}
\label{sec:appendix_limitations}
\ourMethod assumes the visual signature of depth misalignment is readable from the image; for heavily occluded or truncated objects this cue weakens and refinement becomes less reliable, particularly under tight \threeD IoU thresholds. Future work could combine visual alignment with complementary cues such as scene context or temporal consistency. In addition, our projection $\pi(\mathbf{b};\mathbf{K})$ assumes a pinhole camera; extending the alignment view to omnidirectional $360^\circ$ imagery, which is increasingly common in immersive video applications~\cite{zhang2022vrformer, zhang2023tile}, is another natural direction. Since the alignment cue can be subtle for small or distant objects, finer tokenization of discriminative image regions~\cite{zhu2026subtoken} may further sharpen the judgment. Finally, because \ourMethod is agnostic to the upstream source, it could be combined with per-sample selection or fusion of multiple upstream detectors and depth experts before refinement, as recent work does for universal depth estimation~\cite{zhu2026depthagent} and for human recognition~\cite{zhu2025quality, zhu2026fusionagent}.

\section{More Visualization}
\label{sec:appendix-visualization}

\begin{figure}[t]
    \centering
    \includegraphics[width=\linewidth]{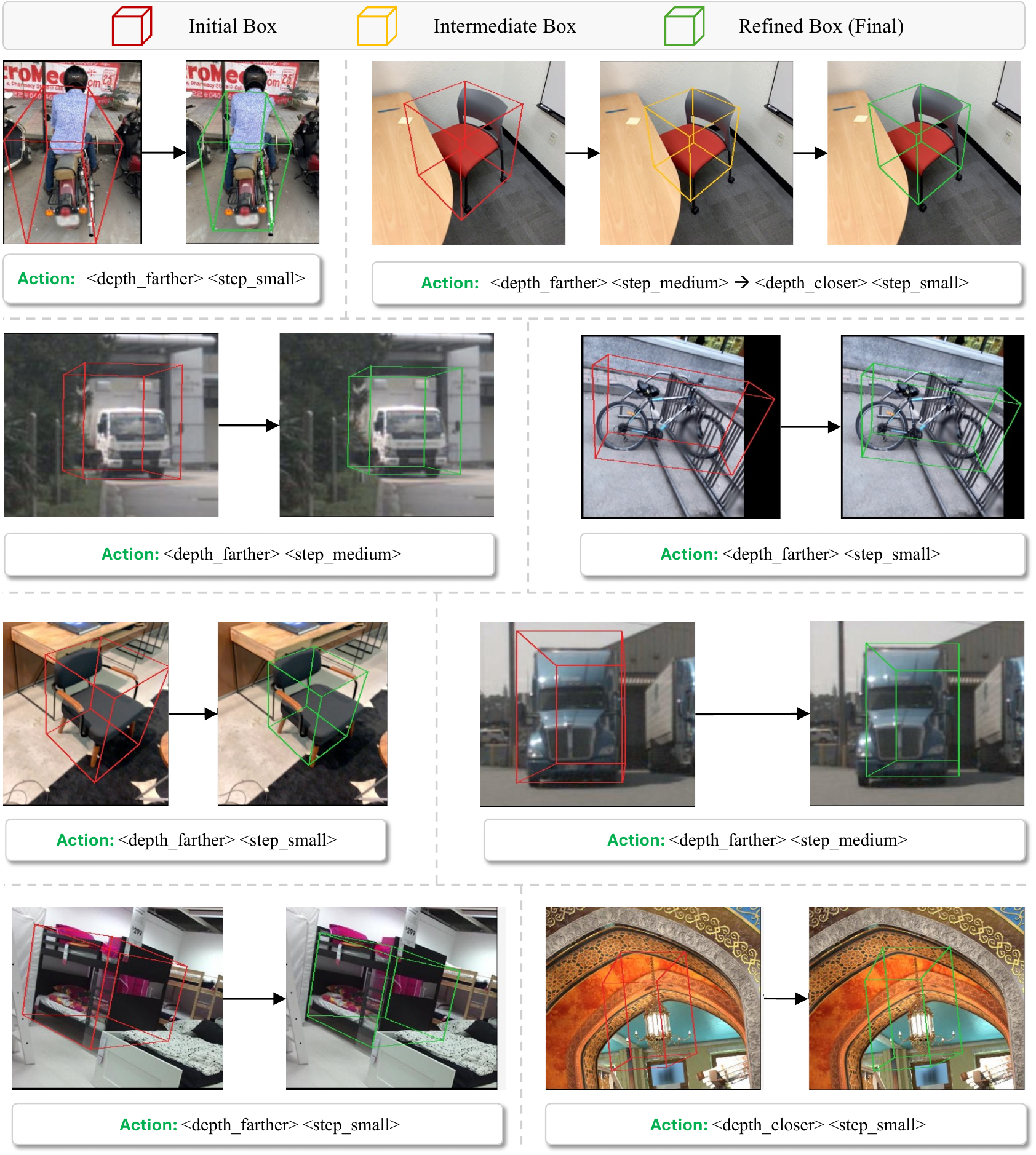}
    \small
    \caption{
        \textbf{Additional qualitative examples of single-step and iterative depth refinement on \omniThreeD.}
        \ourMethod refines DetAny3D~\cite{zhang2025detect}'s initial prediction (red) toward the refined result (green). Most cases are corrected in a single step; harder cases require multiple action steps until the projected box aligns with the target.
    }
    \label{fig:more-vis}
\end{figure}

We provide additional qualitative examples of \ourMethod's refinement behavior in \cref{fig:more-vis}, complementing the main-text examples in \cref{fig:vis}. The examples span the six \omniThreeD~\cite{brazil2023omni3d} sub-datasets and cover diverse object categories, scene types, and initial misalignment patterns. In each row, the leftmost column shows DetAny3D~\cite{zhang2025detect}'s initial prediction (red wireframe) overlaid on the input image, and the rightmost column shows \ourMethod's refined result (green wireframe). For cases that converge in more than one step, the intermediate columns show the projection after each successive action token.

Across these examples, two qualitative patterns are visible. First, the projection tightens around the target object monotonically over iterations: each step either commits to $\langle\texttt{depth\_ok}\rangle$ when the alignment is judged sufficient or moves the projection in the visually correct direction (closer or farther). Second, harder cases tend to involve objects with weak alignment cues, such as small objects, occlusion, or unusual viewpoints, which is where multiple steps are most useful: the first step provides a coarse correction, and subsequent steps make finer adjustments once the projection is roughly in place.

\section{Broader Impact}
\label{sec:appendix_impact}
\ourMethod is a refinement module for monocular \threeD object detection, a perception capability with broad applications in autonomous driving, robotics, augmented reality, and assistive technologies. By improving the object-level depth precision of existing \threeD detectors, \ourMethod can make downstream perception systems safer and more reliable. In autonomous driving, more accurate object localization translates directly into better trajectory prediction and collision avoidance, particularly in the long-tail open-vocabulary regime where current detectors are weakest. In indoor robotics and AR, more precise \threeD geometry enables tighter manipulation and more stable scene understanding, which are critical for assistive applications such as service robots for elderly care or visually impaired users. \ourMethod's plug-and-play design lowers the barrier to adopting these improvements, since practitioners can integrate it on top of existing detectors without retraining the underlying perception stack. Since \ourMethod is fine-tuned from an open-weight VLM, its lineage to the base model can in principle be traced from the weights alone~\cite{chen2026who}, which supports attribution and license compliance when such checkpoints are redistributed. At the same time, more accurate \threeD localization from a single camera could also be used in surveillance or tracking systems without the consent of the people being observed, and a refined box should not be treated as ground truth in safety-critical settings: \ourMethod can still fail on heavily occluded or truncated objects, so deployed systems should keep independent safety checks rather than rely on refined depth alone.

\clearpage
\section*{NeurIPS Paper Checklist}

\begin{enumerate}

\item {\bf Claims}
    \item[] Question: Do the main claims made in the abstract and introduction accurately reflect the paper's contributions and scope?
    \item[] Answer: \answerYes{}
    \item[] Justification: The abstract and \cref{sec:intro} clearly state our four contributions: formalizing \threeD box refinement as a stand-alone task (\cref{sec:intro}), recasting it as semantic alignment (\cref{sec:approach}), instantiating it with action tokens (\cref{sec:approach}), and demonstrating consistent improvements as a plug-and-play module (\cref{sec:experiments}). All claims are supported by the experimental results in \cref{sec:experiments}.

\item {\bf Limitations}
    \item[] Question: Does the paper discuss the limitations of the work performed by the authors?
    \item[] Answer: \answerYes{}
    \item[] Justification: \cref{sec:appendix_limitations} discusses the main limitations: the visual cue weakens for heavily occluded or truncated objects, and the projection assumes a pinhole camera. In addition, \cref{sec:exp_benchmark} acknowledges that performance degrades more under Novel Category than Novel Camera shifts, and \cref{sec:ablation} reports honest comparisons including cases where alternative designs achieve comparable results on certain metrics.

\item {\bf Theory assumptions and proofs}
    \item[] Question: For each theoretical result, does the paper provide the full set of assumptions and a complete (and correct) proof?
    \item[] Answer: \answerNA{}
    \item[] Justification: The paper does not include formal theoretical results requiring proofs. The factorization in \cref{sec:approach} is a standard application of the chain rule for autoregressive sequence modeling.

\item {\bf Experimental result reproducibility}
    \item[] Question: Does the paper fully disclose all the information needed to reproduce the main experimental results of the paper to the extent that it affects the main claims and/or conclusions of the paper (regardless of whether the code and data are provided or not)?
    \item[] Answer: \answerYes{}
    \item[] Justification: \cref{sec:action_tokens,sec:training_inference} fully describe the architecture and training procedure. \cref{app:appendix_impl} provides detailed implementation specifications including hyperparameters, optimizer settings, and hardware.  \cref{app:appendix_benchmark} documents the \benchmarkName benchmark construction.

\item {\bf Open access to data and code}
    \item[] Question: Does the paper provide open access to the data and code, with sufficient instructions to faithfully reproduce the main experimental results, as described in supplemental material?
    \item[] Answer: \answerYes{}
    \item[] Justification: We release code, model checkpoints, and the curated training and benchmark data, together with instructions to reproduce the main results, at \url{https://github.com/alanzhangcs/RefineAny3D}.

\item {\bf Experimental setting/details}
    \item[] Question: Does the paper specify all the training and test details (e.g., data splits, hyperparameters, how they were chosen, type of optimizer) necessary to understand the results?
    \item[] Answer: \answerYes{}
    \item[] Justification: \cref{sec:training_inference} describes the two-stage training procedure. \cref{app:appendix_impl} reports detailed hyperparameters: learning rates ($5\!\times\!10^{-3}$ for Stage 1, $1\!\times\!10^{-5}$ for Stage 2), AdamW optimizer with cosine schedule, gradient clipping, batch size, and training duration. Data splits are specified in \cref{sec:experiments,app:appendix_benchmark}.

\item {\bf Experiment statistical significance}
    \item[] Question: Does the paper report error bars suitably and correctly defined or other appropriate information about the statistical significance of the experiments?
    \item[] Answer: \answerNo{}
    \item[] Justification: We do not report error bars due to the high computational cost of training the full model (approximately $90$ hours on $8$ H100 GPUs per Stage 2 run). We mitigate variability by evaluating on large stratified test sets that aggregate over many samples, and by reporting consistent improvements across multiple datasets and settings.

\item {\bf Experiments compute resources}
    \item[] Question: For each experiment, does the paper provide sufficient information on the computer resources (type of compute workers, memory, time of execution) needed to reproduce the experiments?
    \item[] Answer: \answerYes{}
    \item[] Justification: \cref{app:appendix_impl} reports that all training uses $8$ NVIDIA H100 ($80$~GB) GPUs with bfloat16 precision, DeepSpeed ZeRO-3, and FlashAttention-2. Stage 1 takes approximately $1$ hour and Stage 2 approximately $90$ hours. Inference latency is measured on a single H100 (\cref{sec:appendix-inference}).

\item {\bf Code of ethics}
    \item[] Question: Does the research conducted in the paper conform, in every respect, with the NeurIPS Code of Ethics?
    \item[] Answer: \answerYes{}
    \item[] Justification: The research conforms to the NeurIPS Code of Ethics. We use only publicly available datasets (Omni3D and its source datasets) and pretrained models (Qwen3-VL and Qwen3) under their respective licenses, and our work does not involve human subjects or sensitive personal data.

\item {\bf Broader impacts}
    \item[] Question: Does the paper discuss both potential positive societal impacts and negative societal impacts of the work performed?
    \item[] Answer: \answerYes{}
    \item[] Justification: \cref{sec:appendix_impact} discusses both. Our method improves monocular \threeD perception, which has positive applications in autonomous driving, robotics, and assistive technologies. Potential negative impacts include misuse for surveillance applications. As the method refines existing \threeD detectors rather than introducing new perception capabilities, the marginal risk over existing detection systems is limited.

\item {\bf Safeguards}
    \item[] Question: Does the paper describe safeguards that have been put in place for responsible release of data or models that have a high risk for misuse?
    \item[] Answer: \answerNA{}
    \item[] Justification: Our model and curated data pose no high risk for misuse beyond what is already enabled by existing publicly available \threeD detection systems and pretrained VLMs. The training data is derived from established academic datasets that are already publicly available under their respective licenses.

\item {\bf Licenses for existing assets}
    \item[] Question: Are the creators or original owners of assets (e.g., code, data, models), used in the paper, properly credited and are the license and terms of use explicitly mentioned and properly respected?
    \item[] Answer: \answerYes{}
    \item[] Justification: All datasets (Omni3D~\cite{brazil2023omni3d}, KITTI~\cite{geiger2012we}, nuScenes~\cite{caesar2020nuscenes}) and pretrained models (Qwen3-VL~\cite{Qwen3-VL}) used in the paper are properly cited. Each asset is used in accordance with its respective license.

\item {\bf New assets}
    \item[] Question: Are new assets introduced in the paper well documented and is the documentation provided alongside the assets?
    \item[] Answer: \answerYes{}
    \item[] Justification: We introduce two new assets: (i) a curated training dataset of $\sim\!3$M depth-perturbation samples with chain-of-thought annotations; and (ii) the \benchmarkName benchmark, documented in \cref{sec:appendix_data} and \cref{app:appendix_benchmark}, respectively. Both are released with documentation at \url{https://github.com/alanzhangcs/RefineAny3D}.

\item {\bf Crowdsourcing and research with human subjects}
    \item[] Question: For crowdsourcing experiments and research with human subjects, does the paper include the full text of instructions given to participants and screenshots, if applicable, as well as details about compensation (if any)?
    \item[] Answer: \answerNA{}
    \item[] Justification: The paper does not involve crowdsourcing or human subjects research. All annotations used in training are derived programmatically from existing public \threeD datasets.

\item {\bf Institutional review board (IRB) approvals or equivalent for research with human subjects}
    \item[] Question: Does the paper describe potential risks incurred by study participants, whether such risks were disclosed to the subjects, and whether Institutional Review Board (IRB) approvals (or an equivalent approval/review based on the requirements of your country or institution) were obtained?
    \item[] Answer: \answerNA{}
    \item[] Justification: The paper does not involve research with human subjects.

\item {\bf Declaration of LLM usage}
    \item[] Question: Does the paper describe the usage of LLMs if it is an important, original, or non-standard component of the core methods in this research?
    \item[] Answer: \answerYes{}
    \item[] Justification: LLMs are core components of the method: \ourMethod fine-tunes Qwen3-VL-8B~\cite{Qwen3-VL}, and Qwen3-VL and Qwen3~\cite{qwen3} are used for annotation filtering and chain-of-thought synthesis (\cref{sec:data_curation,sec:appendix_data}). We additionally use an LLM for grammar checking.

\end{enumerate}

\end{document}